%% file: example_paper.tex
\documentclass{article}

\usepackage{microtype}
\usepackage{graphicx}
\usepackage{subcaption}
\usepackage{booktabs} 

\usepackage{hyperref}

\usepackage[accepted]{icml2026}

\usepackage{amsmath}
\usepackage{amssymb}
\usepackage{mathtools}
\usepackage{amsthm}

\usepackage[capitalize,noabbrev]{cleveref}

\theoremstyle{plain}

\theoremstyle{definition}

\theoremstyle{remark}

\usepackage[textsize=tiny]{todonotes}

\usepackage{multirow}
\usepackage[table]{xcolor}  
\usepackage{tabularx}
\usepackage{wrapfig}
\usepackage{xspace}
\usepackage{pifont}
\usepackage{stfloats}
\usepackage{graphicx}
\usepackage{subcaption}

\newcommand{\name}{GeoThinker\xspace}

\newcommand{\module}{Spatial-Grounded Fusion\xspace}
\newcommand{\submodulea}{Frame-Wise Constraints\xspace}
\newcommand{\submoduleb}{Importance Gating\xspace}

\newcommand{\modulelow}{spatial-grounded fusion\xspace}
\newcommand{\submodulealow}{frame-wise constraints\xspace}
\newcommand{\submoduleblow}{importance gating\xspace}

\newcommand{\sota}{\textbf}
\definecolor{navyblue}{HTML}{0071BC}

\definecolor{roborefgreen}{rgb}{0.92, 1.0, 0.92}
\definecolor{roborefgray}{gray}{0.95}
\definecolor{roborefblue}{rgb}{0.88,0.98,1}
\definecolor{roborefred}{rgb}{1, 0.9, 0.9}

\usepackage{xspace}
\makeatletter
\DeclareRobustCommand\onedot{\futurelet\@let@token\@onedot}
\def\@onedot{\ifx\@let@token.\else.\null\fi\xspace}

\makeatother

\newcommand{\cmark}{\ding{51}}%
\newcommand{\cmarkg}{\textcolor{lightgray}{\ding{51}}}%
\newcommand{\xmark}{\ding{55}}%
\newcommand{\xmarkg}{\textcolor{lightgray}{\ding{55}}}%

\begin{document}


\twocolumn[
  \icmltitle{
  Thinking with Geometry: Active Geometry Integration for Spatial Reasoning 
  }




  \icmlsetsymbol{equal}{*}
  \icmlsetsymbol{partner}{$\dag$}

  \begin{icmlauthorlist}
    \icmlauthor{Haoyuan Li}{equal,sysu,partner}
    \icmlauthor{Qihang Cao}{equal,sjtu,partner}
    \icmlauthor{Tao Tang}{sysu}
    \icmlauthor{Kun Xiang}{sysu}
    \icmlauthor{Zihan Guo}{sysu,cz}\\
    \icmlauthor{Jianhua Han}{yinwang}
    \icmlauthor{JiaWang Bian}{ntu}
    \icmlauthor{Hang Xu}{yinwang}
    \icmlauthor{Xiaodan Liang}{sysu,mbz}
  \end{icmlauthorlist}

  \icmlaffiliation{sysu}{Shenzhen campus of Sun Yet-sen University}
  \icmlaffiliation{yinwang}{Yinwang Intelligent Technology Co. Ltd.}
  \icmlaffiliation{mbz}{MBZUAI}
  \icmlaffiliation{sjtu}{Shanghai Jiao Tong University}
  \icmlaffiliation{cz}{Shanghai innovation institute}
  \icmlaffiliation{ntu}{Nanyang Technological University}

  \icmlcorrespondingauthor{Xiaodan Liang}{xdliang328@gmail.com}

  \icmlkeywords{Machine Learning, ICML}

  \vskip 0.1in
  
  {
      \begin{center}
        \centering
        \includegraphics[width=\textwidth]{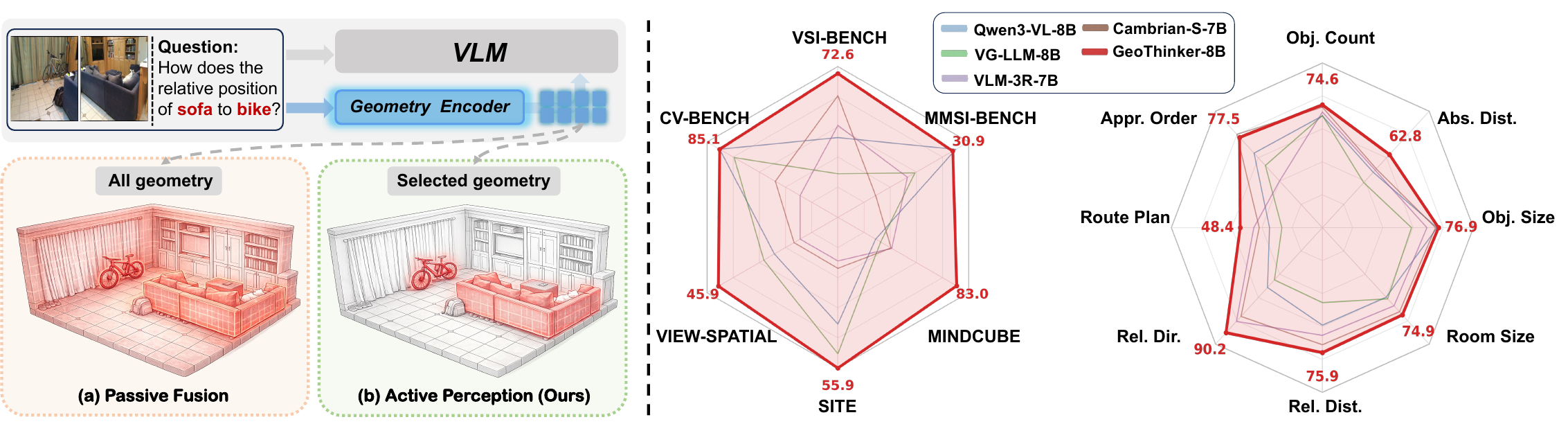}
        \captionof{figure}{
        \textbf{Thinking with geometry through active integration.} \textbf{Left: }\textbf{(a) Passive Fusion:} Conventional MLLMs indiscriminately incorporate a global stream of geometric features, which leads to significant information redundancy and semantic-texture misalignment. \textbf{(b) Active Perception (\name):} Our framework shifts the paradigm by empowering the model to discern and selectively retrieve spatial cues guided by its internal reasoning demands. \textbf{Right:} Active perception yields superior performance across diverse spatial intelligence benchmarks.
        }
        \vspace{-1mm}
      \end{center}
  }

  \vskip 0.1in 
]

\printAffiliationsAndNotice{\icmlEqualContribution and \textsuperscript{$\dag$} Work done as an intern at Yinwang.}

\begin{abstract}

\vspace{-1mm}

Recent progress in spatial reasoning with Multimodal Large Language Models (MLLMs) increasingly leverages geometric priors from 3D encoders. However, most existing integration strategies remain passive: geometry is exposed as a global stream and fused in an indiscriminate manner, which often induces semantic-geometry misalignment and redundant signals.
We propose \emph{\name}, a framework that shifts the paradigm from passive fusion to active perception. 
Instead of feature mixing, \name enables the model to selectively retrieve geometric evidence conditioned on its internal reasoning demands.
\name achieves this through \textit{Spatial-Grounded Fusion} applied at carefully selected VLM layers, where semantic visual priors selectively query and integrate task-relevant geometry via frame-strict cross-attention, further calibrated by \textit{Importance Gating} that biases per-frame attention toward task-relevant structures.
Comprehensive evaluation results show that \name sets a new state-of-the-art in spatial intelligence, achieving a peak score of 72.6 on the VSI-Bench. Furthermore, \name demonstrates robust generalization and significantly improved spatial perception across complex downstream scenarios, including embodied referring and autonomous driving. Our results indicate that the ability to actively integrate spatial structures is essential for next-generation spatial intelligence. Code can be found at \href{https://github.com/Li-Hao-yuan/GeoThinker}{https://github.com/Li-Hao-yuan/GeoThinker}.

\end{abstract}

\input{section/01_introduction}

\input{section/02_related_works}

\input{section/03_method}

\input{section/04_experiments}

\input{section/05_conclusion}

{
    \bibliography{example_paper}
    \bibliographystyle{icml2026}
}

\appendix
\input{section/appendix}

\end{document}

%% file: section/01_introduction.tex
\vspace{-9mm}
\section{Introduction}
\vspace{-1mm}

The pursuit of spatial intelligence has emerged as a pivotal frontier for Multimodal Large Language Models (MLLMs), driving significant advancements in 3D scene understanding~\cite{cai2025scaling(sensenova),yang2025cambrian}, vision-language-action models~\cite{li2025recogdrive,xu2025a0,zhang2024navid}, and embodied intelligence~\cite{zhou2025roborefer,xu20253d}. Central to this evolution is the integration of geometry encoders~\cite{wang2025pi3,wang2025continuous(cut3r)}(e.g., VGGT~\cite{wang2025vggt}), which provide fine-grained spatial priors. These priors enable models to move beyond 2D semantic perception toward a deeper understanding of the structured 3D world.

\begin{figure}[t]
	\centering
        \includegraphics[width=0.48\textwidth]{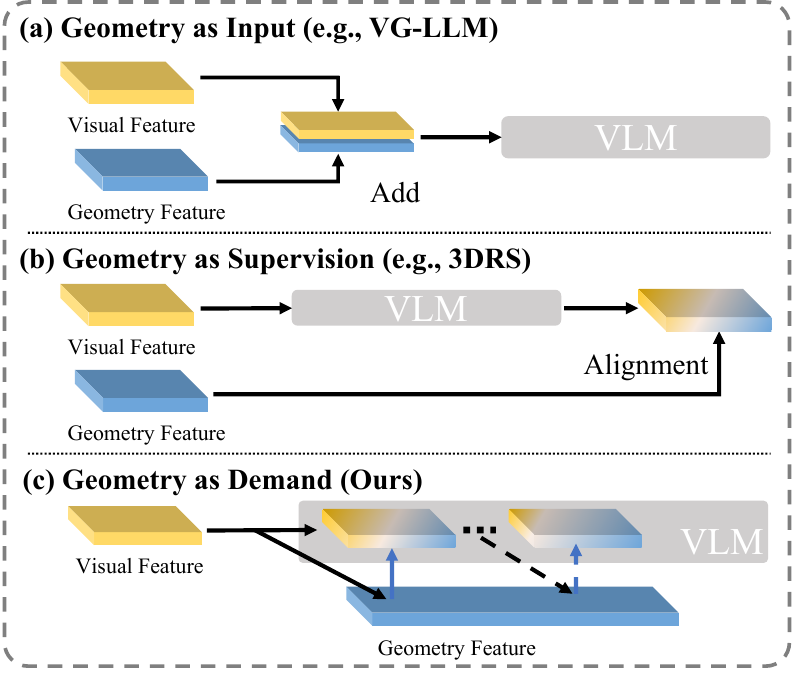}
     \caption{\textbf{Comparison of geometry integration paradigms}. \textbf{(a)} and \textbf{(b)} represent passive paradigms that indiscriminately incorporate geometric streams, often leading to semantic-geometry misalignment and redundant noise. In contrast, \textbf{(c)} \name shifts to active perception, empowering the MLLM to autonomously discern and selectively retrieve task-related geometric cues guided by internal reasoning.}
	\label{fig:motivation}
    \vspace{-7mm}
\end{figure}

Despite these advancements, current geometry integration strategies primarily rely on passive fusion paradigms, as illustrated in \cref{fig:motivation}. Whether through input-level fusion of geometric and semantic features~\cite{zheng2025learning(vgllm),fan2025vlm(vlm3r),chen2025reasoning(gsreasoner),wu2025spatial(spatialmllm)} or geometric knowledge distillation via supervision~\cite{li2025spatialforcing,huang20253drs}, these methods typically treat geometric inputs as a uniformly exposed stream. These one-size-fits-all approaches encounter a critical bottleneck: they overlook the fact that geometric cues are not only task-dependent but also spatially selective. Even for geometry-intensive tasks, the relevant geometric cues are often confined to specific regions of interest rather than the entire scene. Consequently, passive fusion often leads to semantic-geometry misalignment and the injection of redundant noise, which compromises the model's spatial reasoning performance and generalization in complex environments.

To address these challenges, we introduce \name, a framework that shifts the paradigm from passive fusion to active perception. Instead of passively ingesting an indiscriminate geometric stream, \name empowers the MLLM to autonomously discern and retrieve geometric cues based on its internal reasoning demands. The core of \name is Spatial-Grounded Fusion, where semantic visual priors serve as an active bridge to query and fuse task-relevant geometry via frame-strict cross-attention. By constraining attention within each frame, we preserve spatial correspondence between semantic and geometric tokens and prevent cross-frame feature interference. 
In addition, \name incorporates an Importance Gating module that learns a semantic-guided bias over per-frame attention maps, emphasizing task-relevant geometric features (e.g., object boundaries and relational links). Finally, \name applies Spatial-Grounded Fusion at carefully selected layers of the VLM, realizing active perception that mitigates semantic–geometry misalignment and redundant noise.

Extensive experiments show that \name delivers strong and consistent gains across multiple spatial intelligence benchmarks compared with baselines. In particular, \name sets a new state of the art on VSI-Bench, reaching a peak score of 72.6. Under debiased evaluation that reduces non-visual shortcuts, \name remains robust, achieving 68.1 on VSI-Debiased when evaluated with 128-frame video inputs. \name further transfers effectively to demanding downstream settings, improving average accuracy by +1.66\% on embodied referring and boosting PDMS by +2.0 points for autonomous driving.
Collectively, these results suggest that active, semantic-driven integration is a vital step toward building MLLMs with stronger spatial reasoning and a more structured understanding of the 3D world.
\vspace{-1mm}
Our contributions can be summarized as follows:
\begin{itemize}
    \item \textbf{Active perception driven by internal demands.} We propose \name, which enables MLLMs to actively retrieve and integrate geometry conditioned on their internal reasoning needs, rather than passively fusing a uniformly exposed geometry stream.
    \item \textbf{State-of-the-art spatial reasoning performance.} \name achieves SOTA results on spatial intelligence benchmarks, notably best score on VSI-Bench.
    \item \textbf{Robust generalization.} \name remains robust under debiased and long-video evaluation settings, and transfers effectively to diverse downstream scenarios such as embodied referring and autonomous driving.
\end{itemize}

%% file: section/02_related_works.tex
\vspace{-3mm}
\section{Related Work}
\subsection{MLLMs for Spatial Intelligence}
MLLMs\cite{yang2025qwen3technicalreport,team2023gemini,openai_gpt5_systemcard} have achieved impressive progress on general image and video understanding, yet recent benchmarks\cite{yang2025thinking(vsi)} reveal a persistent gap in reliable spatial reasoning, making spatial intelligence a key bottleneck toward human-level capability. To narrow this gap, prior work explores multiple routes. Some methods inject explicit 3D cues into the MLLM pipeline, where Video-3D LLM \cite{zheng2025video(video3dllm)} augments video inputs with per-frame 3D coordinates back-projected from RGB-D to provide position-aware representations. Alternatively, others pursue implicit improvement in latent space: RoSS3D \cite{wang2025ross3d} introduces cross-view and global-view (BEV) reconstruction objectives with denoising-style supervision to encourage geometry-consistent representations. 
In parallel, data scaling has also proven highly effective, Cambrian-S\cite{yang2025cambrian} curates VSI-590K to probe scaling limits, and SenseNova-SI\cite{cai2025scaling(sensenova)} systematically constructs SenseNova-SI-8M to achieve strong gains on VSI-Bench and EASI leaderboard~\cite{cai2025holistic} while maintaining general multimodal capability. 
Complementarily, reasoning-centric training exploits the reasoning capability of LLMs: SpatialLadder\cite{li2025spatialladder} strengthens complex spatial reasoning via reinforcement learning with verifiable rewards, while GS-Reasoner\cite{chen2025reasoning(gsreasoner)} uses grounding-aware CoT supervision to bridge 3D grounding and spatial reasoning. In this work, we focus on efficiently integrating 3D cues from video inputs into MLLMs for improved spatial reasoning.

\begin{figure*}[ht]
	\centering
        \includegraphics[width=0.95\textwidth]{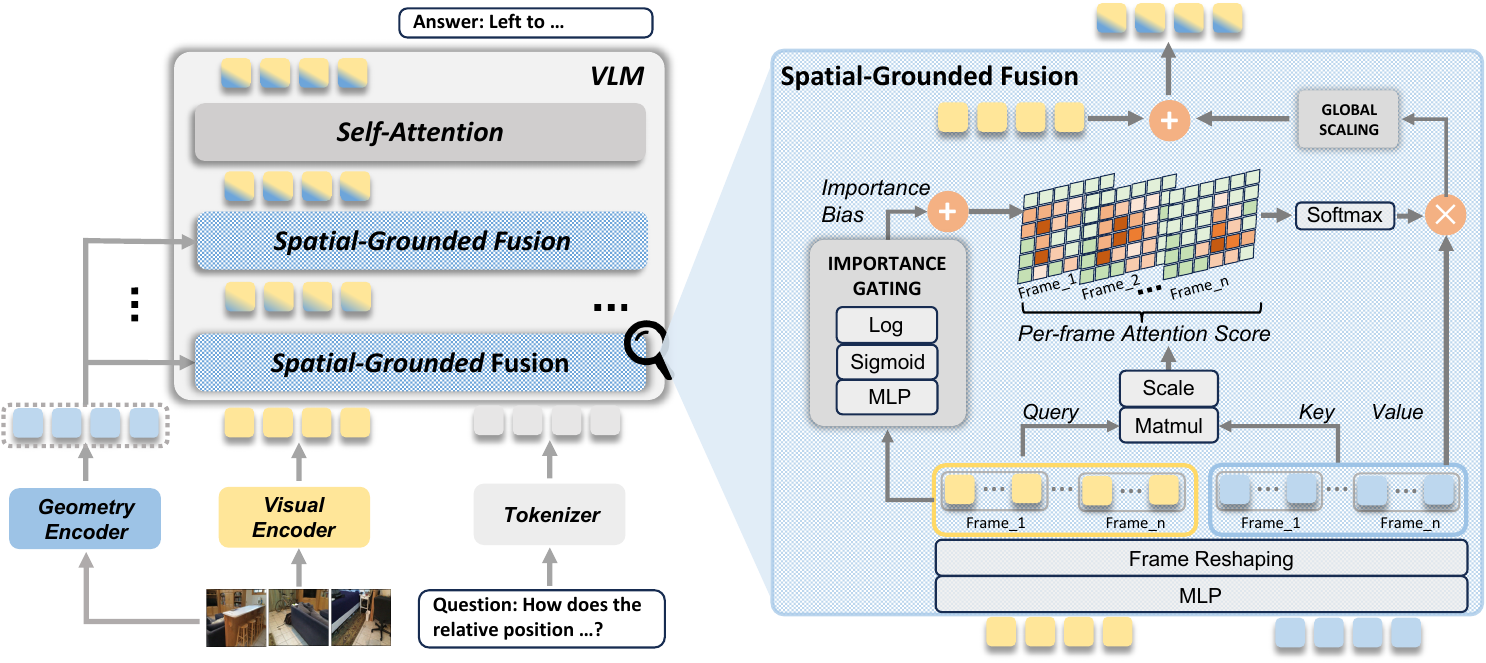}
     \caption{\textbf{Overview of the \name architecture.} Our framework features a decoupled interaction mechanism where the VGGT is integrated via Spatial-Grounded Fusion layers. By employing Importance Gating, the model predicts a localized attention bias to dynamically modulate the injection of geometric textures. This design ensures that rich structural details are only queried when they are contextually relevant to the semantic reasoning process. }
	\label{fig:model_overview}
    \vspace{-5mm}
\end{figure*}

\vspace{-2mm}
\subsection{Geometry-Aware MLLMs}
\vspace{-1mm}
To endow MLLMs with spatial intelligence, recent works begin to incorporate geometry priors from 3D Encoders (e.g., VGGT\cite{wang2025vggt}, $\pi^3$\cite{wang2025pi3}) into Models. Most existing approaches follow passive fusion paradigms. A common practice is input-level fusion, where geometric features are fused with semantic tokens at the model input: VG-LLM\cite{zheng2025learning(vgllm)} performs patch-level addition to form geometry-augmented visual tokens, while VLM-3R\cite{fan2025vlm(vlm3r)} concatenates enriched 3D feature tokens with camera tokens and injects them via cross-attention so visual tokens can query geometry-aware context. Despite the use of cross-attention, geometry remains globally exposed without any retrieval of task-related geometry from noise. As a result, the gap between high-level semantic features and low-level geometry cues can still limit effective integration. 
G$^2$VLM\cite{hu2025g2vlm} proposes a MoT-style architecture with dedicated geometric and semantic experts, jointly learning 3D reconstruction and spatial reasoning through shared self-attention. 
However, it relies on large-scale multi-task training and additional objectives, motivating more efficient geometry integration mechanisms.
In parallel, another line of work adopts feature distillation or alignment. 3DRS\cite{huang20253drs} and 3DThinker\cite{chen2025think(3dthinker)} distills 3D priors from 3D foundation models into MLLM visual representations, while Spatial Forcing\cite{li2025spatialforcing} directly aligns intermediate visual embeddings with geometric representations to enforce spatial structure.
However, these methods inject geometry through training-time supervision, but provide limited control over how geometric evidence is selectively used during inference.
In contrast, our method enables more effective integration by actively selecting task-relevant geometric features conditioned on semantics.

%% file: section/03_method.tex
\input{table/02_VSI_bench}

\vspace{-2mm}
\section{Method}
\vspace{-1mm}

To enhance MLLMs with 3D geometry priors for spatial reasoning, we propose \name, an active integration framework.
As illustrated in \cref{fig:model_overview}, \name shifts the paradigm from passive fusion to active perception. Instead of the indiscriminate feature addition in prior works, we introduce a Spatial-Grounded Fusion (SGF), which allows the MLLM to integrate task-relevant geometric cues conditioned on internal semantic demands via frame-strict cross-attention.
\Cref{sec:method_arch} outlines the overall architecture design. \Cref{sec:method_bridge_feat_fusion} details the Spatial-Grounded Fusion (SGF) module, and \Cref{sec:Layer selection} describes how we deploy SGF in our VLM backbone.

\vspace{-1mm}
\subsection{Architecture}
\label{sec:method_arch}

\textbf{Preliminary.} 
Given a sequence of RGB images $\left\{I_i\right\}_{i=1}^n$ and a natural-language query $Q$, standard Multimodal Large Language Models (MLLMs) typically process a sequence of RGB images by first projecting pixel-level data into a latent visual space. Specifically, a 2D vision encoder maps each image into semantic visual features $T^S_i\in \mathbb{R}^{\left\lfloor\frac{h}{p_s}\right\rfloor \times  \left\lfloor\frac{w}{p_s}\right\rfloor \times c}$, where $I_i\in \mathbb{R}^{h\times w\times 3}$ and $p_s$ is the patch size. These visual tokens are then jointly processed with the text tokens of $Q$ by the LLM for multimodal reasoning and output the response.
In this work, we adopt Qwen-VL series as our foundational backbone. To enhance computational efficiency, Qwen2.5-VL~\cite{Qwen2.5-VL} and Qwen3-VL~\cite{Qwen3-VL} introduce a spatial compression mechanism before LLM layers. Specifically, given the spatial merge size of $2$, it aggregates spatially contiguous $2\times 2$ visual patches into a single representative token, resulting in $T^{S'}_i\in \mathbb{R}^{\left\lfloor\frac{h}{2p_s}\right\rfloor \times  \left\lfloor\frac{w}{2p_s}\right\rfloor \times c}$. This pooling operation significantly reduces the effective sequence length while preserving local semantic integrity, allowing the backbone to efficiently process high-resolution multi-image inputs $T^{S'}$ with natural-language query $Q$.

\textbf{3D Visual Geometry Encoder.} 
To model implicit 3D attributes without explicit 3D supervision, we employ VGGT~\cite{wang2025vggt} as our 3D visual geometry encoder. Unlike vanilla 2D encoders, the visual geometry encoder is designed to understand inter-frame dependencies via a dual-component architecture: an image-wise feature extractor and a cross-frame interaction decoder. 
Let $p_g$ denote the patch size of geometry encoder, we extract the intermediate features $T_i^{G}\in \mathbb{R}^{\left\lfloor\frac{h}{p_g}\right\rfloor \times  \left\lfloor\frac{w}{p_g}\right\rfloor \times c}$ from all input images $\left\{I_i\right\}_{i=1}^n$ jointly, which embed geometry priors necessary for spatial reasoning.
To reconcile the resolution mismatch between the semantic and geometry features, we resample the geometric feature maps to match the token grid used by the MLLM backbone. Since the backbone aggregates spatially adjacent patches into a single token (e.g., a $2\times2$ spatial merge) and $p_g$ may differ from the 2D patch size $p_s$, we interpolate $T_i^{G}$ on a grid aligned with $p_s$ and the merge size of $2$, obtaining $T_i^{G'}\in \mathbb{R}^{\left\lfloor\frac{h}{2p_s}\right\rfloor \times \left\lfloor\frac{w}{2p_s}\right\rfloor \times c}$. This patch-aligned correspondence allows the LLM backbone to query geometric cues at the exact spatial locations aligned with the corresponding semantic regions.

\vspace{-1mm}
\input{table/01_All_Spatial}
\subsection{\module}
\label{sec:method_bridge_feat_fusion}

To overcome the limitations of passive and indiscriminate fusion, we propose Spatial-Grounded Fusion (SGF) for active geometry integration. SGF comprises two key components: Frame-wise Constraints that preserve spatial correspondence and Importance Gating coupled with Global Scaling to prioritize salient geometric cues while filtering redundant noise.

\vspace{-1mm}
\subsubsection{\submodulea}

Within each fusion layer, we facilitate interaction between the image hidden states and geometric cues via a frame-strict cross-attention. Let $\textbf{H}^{img}_{j}\in \mathbb{R}^{(n\times L) \times c}$ denote the hidden states of the image tokens in the $j$-th layer of LLM, where $L=\left\lfloor\frac{h}{2p_s}\right\rfloor \times  \left\lfloor\frac{w}{2p_s}\right\rfloor$ denotes the token length of each image. To preserve spatial consistency and prevent cross-image feature interference, we reshape both $\textbf{H}^{img}_{j}$ and $\textbf{T}^{G'}$ back into their original spatial dimensions, resulting in reshaped features $\textbf{SH}^{img}_{j}\in \mathbb{R}^{n \times L \times c}$ and $\textbf{ST}^{G'}\in \mathbb{R}^{n \times L \times c}$, respectively. And the query, key, and value projections are computed as:
\begin{equation}
    \!\!\!\!\!\textbf{Q}_j\! =\!\! \text{MLP}(\textbf{SH}^{img}_{j}),\!
    \textbf{K}_j\! =\!\! \text{MLP}(\textbf{ST}^{G'}),\!\!
    \textbf{V}_j\! =\!\! \text{MLP}(\textbf{ST}^{G'})\\
\end{equation}
This preserves frame-wise spatial alignment. Specifically, each semantic query attends only to geometric cues from the same frame, maintaining high generalization for multi-view and video inputs.

\subsubsection{\submoduleb}

Recognizing that not all visual regions require geometric cues for reasoning, we introduce Importance Gating to regulate geometry information flow. We predict an importance score $\mathbf{S}_{imp}$ directly from the image hidden states using a lightweight MLP: 
\begin{equation}
    \mathbf{S}^{imp}_{j} = \text{Sigmoid}(\text{MLP}(\mathbf{SH}^{img}_{j})). \\
\end{equation}
We then convert this score into an additive attention bias:
\begin{equation}
    \mathbf{S}^{bias}_{j} = \log(\mathbf{S}^{imp}_{j}+\epsilon). \\
\end{equation}
where $\epsilon$ is a small constant for numerical stability. We add $\mathbf{S}^{bias}_{j}$ to the cross-attention logits to further emphasize task-relevant geometric cues and suppress irrelevant geometry. Therefore, the constrained cross-attention with importance gating can be formulated as:
\begin{equation}
   \text{Attn}(\mathbf{Q}_j, \!\mathbf{K}_j,\! \mathbf{V}_j,\! \mathbf{S}^{bias}_j)\! =\! \text{softmax}(\frac{\textbf{Q}_j\textbf{K}^T_j}{\sqrt{d_k}} \!+\!\mathbf{S}_{j}^{bias})\textbf{V}_j.
\end{equation}

\vspace{-3mm}
\subsubsection{Global Scaling}
\vspace{-1mm}
To control the overall intensity of the geometric injection, we employ a global learnable scalar $\alpha$ for the cross-attention output, which is initialized to 0. Specifically, the fused feature can be calculated as:
\begin{equation}
    \hat{\mathbf{H}}_{j}^{img} = \mathbf{H}_{j}^{img} + \tanh(\alpha) \cdot \text{Attn}(\mathbf{Q}_j, \mathbf{K}_j, \mathbf{V}_j, \mathbf{S}^{bias}_j).
\end{equation} 
The resulting $\hat{\mathbf{H}}_{j}^{img}$ serves as the output of SGF and is added back to the main LLM residual stream. By combining these mechanisms, \name achieves a balance between thinking semantically and querying geometrically, ensuring that geometric information is used precisely and efficiently.

\vspace{-2mm}
\subsection{Layer Selection}
\vspace{-1mm}
\label{sec:Layer selection}

To inject geometry without degrading the backbone’s native semantic understanding, we carefully choose where to apply SGF across layers. We select candidate fusion layers according to a fusion ratio $\rho\in(0,1)$ with boundary constraints to safeguard performance. Concretely, for the Qwen-VL backbone, we first exclude Qwen3-VL’s deep-stacked visual layers~\cite{Qwen3-VL} to avoid perturbing the backbone’s early visual processing. Second, we adopt a configurable start offset: while the model defaults to fusion from the first LLM layer for spatial-centric tasks, we defer fusion for more general-purpose benchmarks, ensuring that subsequent geometric queries are contextually grounded. Finally, we reserve an end buffer by avoiding fusion in the final layers, which helps preserve instruction-following priors and stabilizes response generation. Together, these constraints ensure that geometry acts as an internal reasoning aid rather than a distractor.

%% file: table/02_VSI_bench.tex
\begin{table*}[ht]
  \caption{
  \textbf{Performance comparisons on VSI-Bench (Vanilla regime).} \name outperforms VG-LLM baseline across different model scales, revealing the effectiveness of Spatial-Grounded Fusion.
  }
  \vspace{-3mm}
  \label{tab:vsi-bench-main}
  \begin{center}
    \begin{small}
      \begin{sc}
        \resizebox{\textwidth}{!}{%
        \begin{tabular}{l|c|c|cccc|cccc}
          \toprule
          \multirow{2}{*}{Methods}& Active & \multirow{2}{*}{Avg.}
          & \multicolumn{4}{c|}{\cellcolor{orange!10}Numerical Answer}
          & \multicolumn{4}{c}{\cellcolor{yellow!10}Multiple-Choice Answer} \\
          \cmidrule(lr){4-7}\cmidrule(lr){8-11}
          & perception &&  Obj.Count & Abs. Dist. & Obj. Size & Room Size & Rel. Dist. & Rel. Dir. & Route Plan & Appr. Order  \\
          \midrule
          \rowcolor{black!11}
          \multicolumn{11}{l}{\textbf{Baseline}} \\
          Chance Level (Random)   &   & --   & --   & --   & --   & --   & 25.0   & 36.1   & 28.3   & 25.0 \\
          Chance Level (Frequency)  &  & 34.0 & 62.1 & 32.0 & 29.9 & 33.1 & 25.1 & 47.9 & 28.4 & 25.2 \\
          \midrule
          \rowcolor{black!8}
          \multicolumn{11}{l}
          {\textbf{Proprietary Models (API)}} \\
          GPT-4o\cite{hurst2024gpt4o}            &  & 34.0 & 46.2 & 5.3  & 43.8 & 38.2 & 37.0 & 41.3 & 31.5 & 28.5 \\
          Gemini-1.5 Flash\cite{team2024gemini(gemini1.5)}   &  & 42.1 & 49.8 & 30.8 & 53.5 & 54.4 & 37.7 & 41.0 & 31.5 & 37.8 \\
          Gemini-1.5 Pro\cite{team2024gemini(gemini1.5)}     &  & 45.4 & 56.2 & 30.9 & 64.1 & 43.6 & 51.3 & 46.3 & 36.0 & 34.6 \\
          \midrule
          \rowcolor{black!11}
          \multicolumn{11}{l}{\textbf{Open-sourced Models}} \\
          LLaVA-OneVision-7B\cite{li2024llava-onevision}      &  & 32.4 & 47.7 & 20.2 & 47.4 & 12.3 & 42.5 & 35.2 & 29.4 & 24.4 \\
          LLaVA-OneVision-72B\cite{li2024llava-onevision}     &  & 40.2 & 43.5 & 23.9 & 57.6 & 37.5 & 42.5 & 39.9 & 32.5 & 44.6 \\
          LLaVA-NeXT-Video-7B\cite{liu2024llavanext}     &  & 35.6 & 48.5 & 14.0 & 47.8 & 24.2 & 43.5 & 42.4 & 34.0 & 30.6 \\
          LLaVA-NeXT-Video-72B\cite{liu2024llavanext}  &  & 40.9 & 48.9 & 22.8 & 57.4 & 35.3 & 42.4 & 36.7 & 35.0 & 48.6 \\
          InternVL2-8B\cite{chen2024internvl2}           &  & 34.6 & 23.1 & 28.7 & 48.2 & 39.8 & 36.7 & 30.7 & 29.9 & 39.6 \\
          InternVL2-40B\cite{chen2024internvl2}         &  & 36.0 & 34.9 & 26.9 & 46.5 & 31.8 & 42.1 & 32.2 & 34.0 & 39.6 \\
          
          Qwen2.5VL-3B\cite{Qwen2.5-VL} &  & 28.6 & 32.7 & 19.5 & 17.3 & 25.1 & 37.3 & 44.9 & 30.4 & 21.8 \\
          Qwen2.5VL-7B\cite{Qwen2.5-VL} &  & 29.3 & 25.2 & 10.5 & 36.4 & 29.6 & 38.4 & 38.0 & 29.8 & 26.8 \\
          
          \midrule
          \rowcolor{black!11}
          \multicolumn{11}{l}{\textbf{Open-source Spatial Intelligence Models}} \\

          SPAR-8B\cite{zhang2025flatland(spar)} & \textcolor{lightgray}{--} & 44.1 & -- & -- & -- & -- & -- & -- & -- & -- \\

          SpatialLadder-3B\cite{li2025spatialladder} & \textcolor{lightgray}{--} & 44.8 & -- & -- & -- & -- & -- & -- & -- & -- \\
          
          Spatial-MLLM-4B\cite{wu2025spatial(spatialmllm)} & \xmarkg & 48.4 & 65.3 & 34.8 & 63.1 & 45.1 & 41.3 & 46.2 & 33.5 & 46.3 \\
        
          VG-LLM-4B\cite{zheng2025learning(vgllm)} & \xmarkg & 46.7 & 67.6 & 37.6 & 55.2 & 52.5 & 48.0 & 44.7 & 31.9 & 35.5 \\

          VG-LLM-8B~\cite{zheng2025learning(vgllm)} & \xmarkg & 49.7 & 68.1 & 38.7 & 59.0 & 61.1 &45.5 &44.9 &26.8 &53.4\\

        \midrule
          \rowcolor{navyblue!11}
          \multicolumn{11}{l}{\textbf{Ours}} \\
          
          \name\textsubscript{Qwen2.5VL-3B}  & \cmark & \sota{48.9} & 68.5 & 36.1 & 57.3 & 62.5 & 43.7 & 47.9 & 34.5 & 40.9 \\
          
          \name\textsubscript{Qwen2.5VL-7B} & \cmark & \sota{50.5} & 69.5 & 38.5 & 57.9 & 62.2 & 45.2 & 46.2 & 31.4 & 52.6 \\

          \bottomrule
        \end{tabular}%
        }
      \end{sc}
    \end{small}
  \end{center}
  \vskip -0.2in
\end{table*}

%% file: table/01_All_Spatial.tex
\begin{table*}[ht]
  \caption{\textbf{Cross-benchmark comparison on spatial intelligence benchmarks (Scaled regime).}
  $\dagger$ indicates evaluation on reduced subsets. * indicates trained with S1+S2 dataset setting from VG-LLM. Benchmarks include VSI-Bench\cite{yang2025thinking(vsi)}, MMSI-Bench\cite{yang2025mmsi}, MindCube-Tiny\cite{yin2025spatial(mindcube)}, Viewspatial\cite{li2025viewspatial}, SITE\cite{wang2025site}, CV-Bench~\cite{tong2024cambrian(cambrian-1)}. \xmark ~ denotes passive usage and -- denotes methods without dedicated geometry encoders.
  }
  \vskip -0.1in
  \label{tab:cross-benchmark}
  \begin{center}
    \begin{small}
      \begin{sc}
        \resizebox{\textwidth}{!}{%
        \begin{tabular}{l|c|c|cccccc}
          \toprule

          \multirow{2}{*}{\textbf{Models}}
          & \textbf{Active}
          & \multirow{2}{*}{\textbf{AVG.}}
          & \multirow{2}{*}{\textbf{VSI-Bench}}
          & \multirow{2}{*}{\textbf{MMSI-Bench}}
          & \multirow{2}{*}{\textbf{MindCube}}
          & \multirow{2}{*}{\textbf{ViewSpatial}}
          & \multirow{2}{*}{\textbf{SITE}}
          & \multirow{2}{*}{\textbf{CV-Bench}}\\

          & \textbf{perception} & & & & & & & \\

          \midrule
          Human & & -- & 79.2 & 97.2 & 94.5 & -- & 67.5 & --  \\
          Random Choice & & -- & 34.0 & 25.0 & 33.0 & 26.3 & 0.0 & --  \\
          \midrule
          \rowcolor{black!9}
          \multicolumn{9}{l}{\textbf{Proprietary Models}} \\
          Seed-1.6\cite{seed2025seed1_5vl}
            & & 53.41 & 49.9 & 38.3 & 48.7 & 43.8 & 54.6 & 85.2 \\
          Gemini-2.5-Pro\cite{team2023gemini}
            & & 56.33 & 53.5 & 38.0 & 57.6 & 46.0 & 57.0 & 85.9 \\
          GPT-5\cite{openai_gpt5_systemcard}
            & & 57.50 & 55.0 & 41.8 & 56.3 & 45.5 & 61.8 & 84.6 \\
          Gemini-3-Pro-Preview \cite{gemini_3_pro_systemcard}
            & & 62.16 & 52.5 & 45.2 & 70.8 & 50.3 & 62.2 & 92.0 \\
          \midrule
          \rowcolor{black!9}
          \multicolumn{9}{l}{\textbf{Open-source General Models}} \\
          Bagel-7B-MoT\cite{deng2025bagel}
            & & 41.90 & 31.4 & 31.0 & 34.7 & 41.3 & 37.0 & 76.0\\
          Qwen2.5-VL-3B-Instruct\cite{Qwen2.5-VL}
            & & 38.60 & 28.6 & 28.6 & 37.6 & 31.9 & 33.1 & 71.8 \\
          Qwen2.5-VL-7B-Instruct\cite{Qwen2.5-VL}
            & & 40.31 & 29.3 & 26.8 & 36.0 & 36.8 & 37.6 & 75.4  \\
          Qwen3-VL-2B-Instruct\cite{Qwen3-VL}
            & & 42.40 & 49.4 & 11.9 & 31.4 & 34.2 & 35.6 & 78.4 \\
          Qwen3-VL-8B-Instruct\cite{Qwen3-VL}
            & & 47.70 & 57.7 & 28.8 & 29.8 & 39.0 & 45.8 & 85.1 \\
          InternVL3-2B\cite{zhu2025internvl3}
            & & 39.31 & 32.9 & 26.5 & 37.5 & 32.5 & 30.0 & 76.5 \\
          InternVL3-8B \cite{zhu2025internvl3}
            & & 45.38 & 42.1 & 28.0 & 41.5 & 38.6 & 41.1 & 81.0 \\
          \midrule
          \rowcolor{black!9}
          \multicolumn{9}{l}{\textbf{Open-source Spatial Intelligence Models}} \\
          
          SpatialLadder-3B\cite{li2025spatialladder}
            & \textcolor{lightgray}{--} & 42.83 & 44.8 & 27.4 & 43.4 & 39.8 & 27.9 & 73.7 \\
          VST-3B-SFT \cite{vst2025}
            & \textcolor{lightgray}{--} & 49.50 & 57.9$\dagger$ & 30.2$\dagger$ & 35.9 & 52.8 & 35.8 & 84.4\\
          VST-7B-SFT\cite{vst2025}
           & \textcolor{lightgray}{--} & 51.31 & 60.6$\dagger$ & 32.0$\dagger$ & 39.7 & 50.5 & 39.6 & 85.5   \\
          Cambrian-S-3B\cite{yang2025cambrian}
            & \textcolor{lightgray}{--} & 42.91 & 57.3 & 25.2 & 32.5 & 39.0 & 28.3 & 75.2 \\
          Cambrian-S-7B \cite{yang2025cambrian}
            & \textcolor{lightgray}{--} & 47.28 & 67.5 & 25.8 & 39.6 & 40.9 & 33.0 & 76.9 \\
          SenseNova-SI-1.1\textsubscript{Qwen3VL-8B}~\cite{cai2025scaling(sensenova)} & \textcolor{lightgray}{--} & - & 64.8 & 38.1 & 73.8 & 51.2 & 49.6 & - \\
          SenseNova-SI-1.1\textsubscript{InternVL3-8B}~\cite{cai2025scaling(sensenova)} & \textcolor{lightgray}{--} & - & 68.8 & 43.3 & 85.7 & 54.7 & 47.7 & - \\

          3DThinker*\textsubscript{Qwen2.5-7B}\cite{chen2025think(3dthinker)} & \xmarkg & - & 63.7 & 43.3 & 76.0 & 68.6  & -  & 81.1 \\
          VLM-3R-7B\cite{fan2025vlm(vlm3r)}
           & \xmarkg & 45.40 & 60.9 & 27.9 & 40.0 & 40.5 & 31.3 & 71.8 \\
          VG-LLM-4B\cite{zheng2025learning(vgllm)} & \xmarkg & 47.46 & 46.6 & 28.0 & 36.9 & 42.5  & 49.8  & 81.0\\
          VG-LLM-8B\cite{zheng2025learning(vgllm)} & \xmarkg & 48.15 & 49.6 & 28.4  & 32.7  & 42.9  & 52.6  & 82.7  \\
          VG-LLM-8B*\cite{zheng2025learning(vgllm)} & \xmarkg & 51.05 & 62.2 & 30.0 & 36.1 & 45.8 & 50.5 & 81.7 \\

            \midrule
            \rowcolor{navyblue!9}
            \multicolumn{9}{l}{\textbf{Ours}} \\
            \name\textsubscript{Qwen2.5VL-7B} & \cmark & \sota{60.43} & 68.5 & 31.7 & 83.6 & 41.9 & 54.8 & 82.1 \\
            
            \name\textsubscript{Qwen3VL-8B} & \cmark & \sota{62.23} & 72.6 & 30.9  & 83.0  & 45.9 & 55.9 & 85.1  \\

          \bottomrule
        \end{tabular}}
      \end{sc}
    \end{small}
  \end{center}
  \vspace{-5mm}
\end{table*}

%% file: section/04_experiments.tex
\vspace{-3mm}
\section{Experiments}
\vspace{-1mm}
In this section, we first provide implementation details, followed by evaluation results on spatial reasoning benchmarks in \cref{exp:spatial reasoning}, demonstrating the effectiveness of our approach. We then present downstream evaluations in \cref{exp:downstream} to validate practical generalization. Next, we conduct an ablation study in \cref{exp:ablation} to verify the contribution of each component in GeoThinker. Finally, in \cref{exp:visualization}, we provide visualizations of our core designs to better interpret model behavior.

\textbf{Implementation Details.} 
To better assess the effectiveness and generalization of our design, we test \modulelow across multiple VLM backbones under different training regimes. For spatial reasoning, we adopt three incremental training settings, all using a batch size of 64 and a learning rate of 1e-5. First, following VG-LLM \cite{zheng2025learning(vgllm)}, the training step is set to 4,656 and the fusion ratio $\rho$ is set to 0.5. Next, by scaling up the VSI-Bench instruction data, we increase the training steps to 21,504 and the fusion ratio $\rho$ is set to 0.75. Finally, we further incorporate general video data from \cite{yang2025cambrian}, which brings the total training steps to 28,235.
For downstream scenarios, we conduct spatial-enhanced training on embodied referring and autonomous-driving planning. For RoboRefer~\cite{zhou2025roborefer}, we use 13,456 steps with batch size 384 and learning rate $1\times10^{-3}$. For ReCogDrive~\cite{li2025recogdrive}, we use 15,213 steps with batch size 128 and learning rate $4\times10^{-5}$.
All experiments are conducted on 8 NVIDIA H800 GPUs.

\input{table/07_video_scaling}

\vspace{-2mm}
\subsection{Spatial Reasoning}
\label{exp:spatial reasoning}
\vspace{-1mm}
\subsubsection{Setting}
\vspace{-1mm}
\textbf{Baseline.} VG-LLM~\cite{zheng2025learning(vgllm)} integrates geometry features from VGGT~\cite{wang2025vggt} into MLLMs via input-level fusion, serving as our primary baseline.
\textbf{(1) Vanilla regime:} Following the VG-LLM configuration, we utilize sampled subsets from SPAR-7M \cite{zhang2025flatland(spar)} and the LLaVA-Hound split of LLaVA-Video-178K \cite{zhang2024video} for fine-tuning.We uniformly sample 8 frames per scene for consistency with the baseline.
\textbf{(2) Scaled regime:} To probe the performance ceiling, we scale the training set with data from VLM-3R \cite{fan2025vlm(vlm3r)}, VSI-590K \cite{yang2025cambrian}, PhysGame \cite{cao2024physgame}, and MindCube \cite{yin2025spatial(mindcube)}. 
We increase the sampling density to 32 frames per scene, and additionally incorporate 430k general video samples from~\cite{yang2025cambrian} to strengthen video understanding.
\subsubsection{Evaluation Results}
We conduct evaluation across multiple benchmarks, including VSI-Bench~\cite{yang2025thinking(vsi)}, MMSI-Bench~\cite{yang2025mmsi}, MindCube~\cite{yin2025spatial(mindcube)}, VideSpatial~\cite{li2025viewspatial}, SITE~\cite{wang2025site}, and CV-Bench~\cite{tong2024cambrian(cambrian-1)}.

\textbf{Vanilla regime:} To evaluate the generalization of our method, we conduct experiments on the VSI-Bench following the evaluation protocol established by VG-LLM. For fair comparison, we keep the same backbone and encoders (Qwen2.5-VL, SigLIP, and VGGT) and only modify the model design.
As shown in \cref{tab:vsi-bench-main}, our \name consistently outperforms VG-LLM across both 3B and 7B scales, achieving higher average scores of 48.9 and 50.5, respectively. This performance gain suggests that our proposed \modulelow is more effective than conventional input-level fusion, by selectively emphasizing task-relevant regions instead of uniformly injecting all geometry.

\input{table/04_VSI_Debiased}

\textbf{Scaled regime:} To evaluate how performance scales with training data, we expand the training mixture by adding VSI-Bench spatial-reasoning instructions and large-scale general video data. To ensure that the model develops generalized spatial reasoning capabilities rather than overfitting to a single benchmark, we evaluate it across a diverse set of tasks and focus on the average performance as the primary metric. As illustrated in \cref{tab:cross-benchmark}, our \name achieves state-of-the-art performance, outperforming both specialized general and specialized spatial models and leading proprietary models. Specifically, our \name\textsubscript{Qwen2.5VL-7B} and \name\textsubscript{Qwen3VL-8B} variant reaches a peak AVG. of 60.43 and 62.23 respectively, demonstrating a comprehensive and balanced mastery of spatial-temporal understanding. 

\noindent \textbf{Robustness to general-video mixture.}
To assess whether scaling with general video data interferes with spatial reasoning, we mix in general video data during training and compare it with the state-of-the-art Cambrian-S-7B. As shown in \cref{tab:comparisons_video_scaling}, Cambrian-S-7B exhibits a clear trade-off: adding 3M general video samples improves temporal benchmarks, but reduces VSI-Bench by 4.1 points (69.2 $\rightarrow$ 65.1). We attribute this to the inherent sensitivity of pure 2D VLM frameworks to data distribution. The infusion of large-scale general video data often disrupts the fine-grained spatial representations required by VSI-Bench.

In contrast, \name benefits from adding general video data without sacrificing VSI-Bench performance. 
With a smaller data mixture of 430k samples, \name not only achieves +5.7 and +26.3 gains on VideoMME and MVBench respectively, but also maintains and even slightly improves its VSI-Bench performance by +0.6. This suggests that \name effectively mitigates task interference: it can selectively leverage geometric cues for spatial reasoning while retaining strong temporal understanding, leading to more robust representations than standard architectures.


\input{table/03_RefSpatial}

\input{table/05_recogdrive}

\noindent \textbf{Robustness against language bias and frame ablation.} To investigate whether our model genuinely relies on visual cues rather than linguistic priors~\cite{li2025does}, we evaluate its performance on the VSI-Debiased benchmark~\cite{brown2025benchmark}. As reported in \cref{tab:comparisons_vsidebiased}, our \name consistently outperforms existing state-of-the-art models, such as Cambrian-S-7B~\cite{yang2025cambrian} and VG-LLM-8B~\cite{zheng2025learning(vgllm)}, across both standard~\cite{yang2025thinking(vsi)} and debiased settings~\cite{brown2025benchmark}. 
Moreover, despite being trained with at most 8/32 frames per sample, \name generalizes to longer contexts at inference: \name\textsubscript{Qwen3VL-8B-32frame} reaches 68.1 on VSI-Debiased with 128 frames, surpassing Cambrian-S-7B (59.9), which is trained with 128-frame windows. This consistent lead on debiased benchmarks confirms that our superior performance stems from a robust spatial understanding rather than over-reliance on language shortcuts.

\vspace{-1mm}
\subsection{Downstream Scenarios}
\label{exp:downstream}

\subsubsection{Embodied Referring}

\textbf{Baseline.} RoboRefer~\cite{zhou2025roborefer} is designed for embodied spatial referring. 
Following its pipeline, we apply the official depth-alignment recipe and then incorporate \modulelow with VGGT into fine-tuning stage. 
We evaluate on RefSpatial-Bench~\cite{zhou2025roborefer}.

\noindent \textbf{Results.} We evaluate our proposed \modulelow with RoboRefer on the challenging RefSpatial-Bench, which contains three splits: location, placement, and unseen compositional spatial relation. As reported in \cref{tab:comparison_roborefer}, \name improves performance on all splits. Compared with the RoboRefer baseline~\cite{zhou2025roborefer}, \name yields +1.00\% on \textit{location} (48.00\% vs. 47.00\%), +1.00\% on \textit{placement} (47.00\% vs. 46.00\%), and +3.89\% on \textit{unseen}, resulting in a +1.66 gain in Avg. Acc. The gains on \textit{location} and \textit{placement} suggest more accurate geometry-aware grounding, which demonstrate effectiveness of \modulelow. While the larger improvement on \textit{unseen} indicates stronger compositional generalization of our \modulelow to novel spatial relations.


\begin{figure*}[t!]
     \centering

     \begin{subfigure}[b]{0.49\textwidth}
         \centering
         \includegraphics[width=\textwidth]{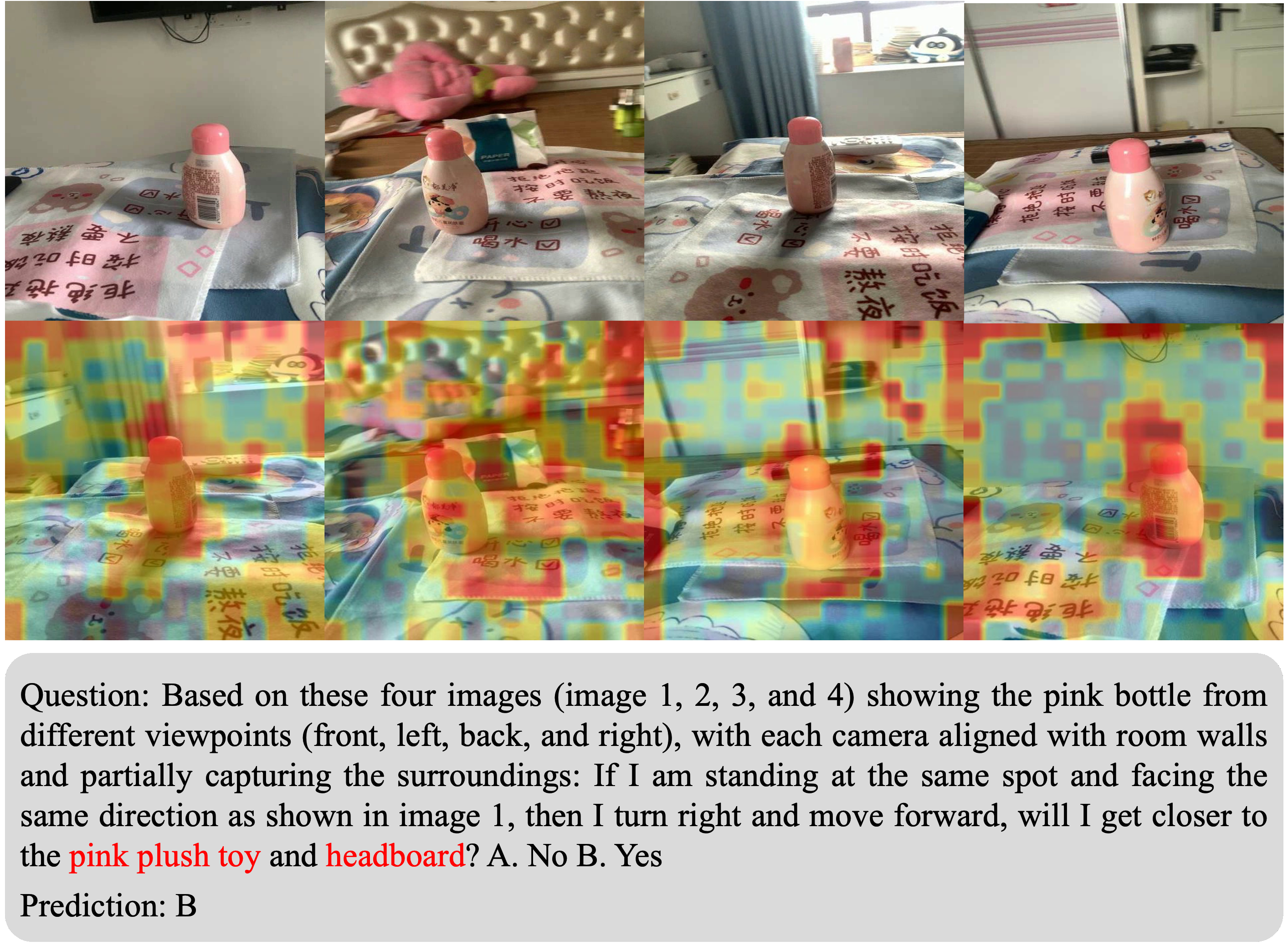}
         \label{fig:appendix_mindube_visual_a}
     \end{subfigure}
     \hfill 
     \begin{subfigure}[b]{0.49\textwidth}
         \centering
         \includegraphics[width=\textwidth]{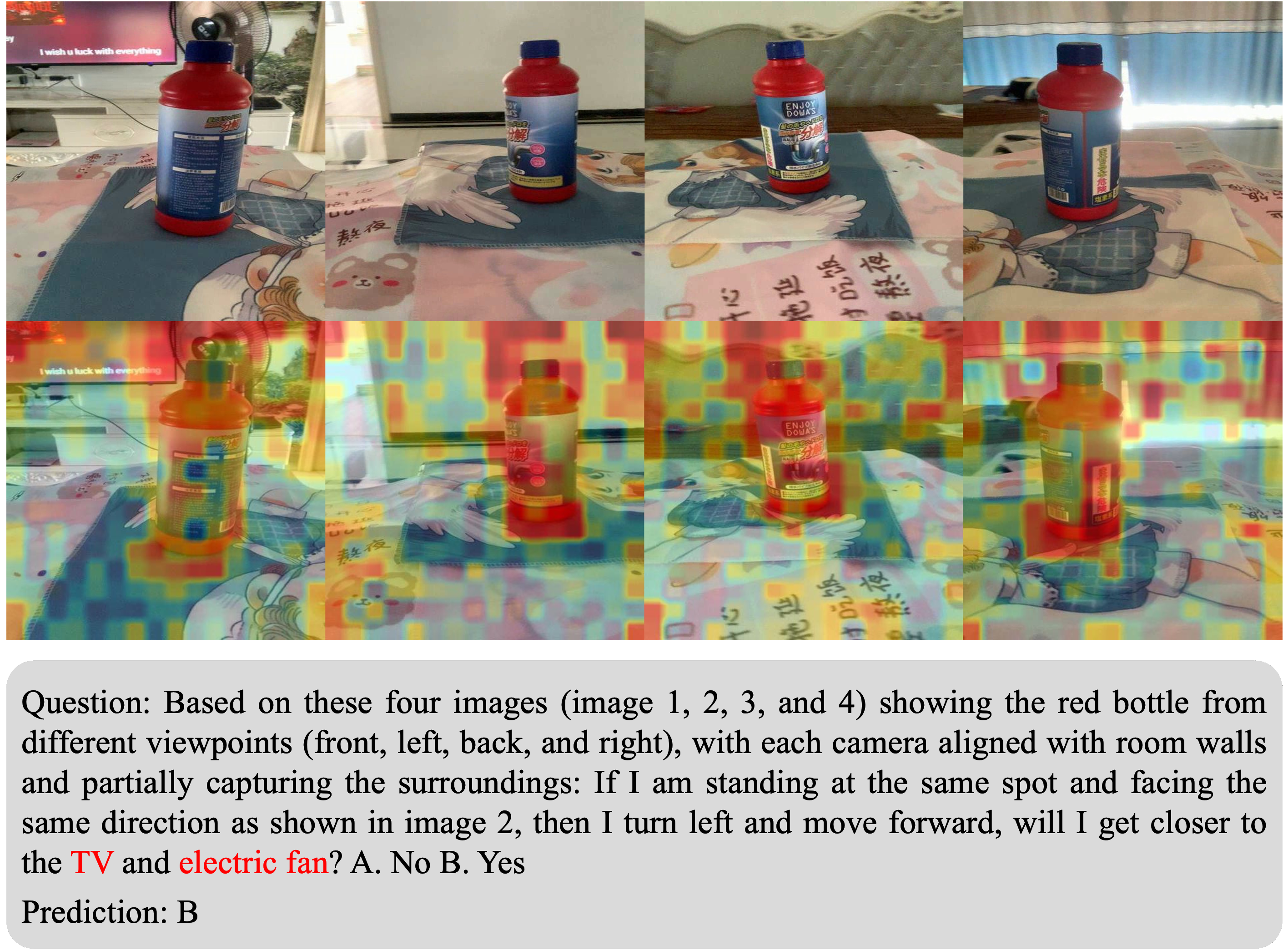}
         \label{fig:appendix_mindube_visual_b}
     \end{subfigure}
     \vspace{-3mm}
     \caption{\textbf{Visualization of Importance Gating Scores.} Heatmaps illustrate that \name naturally learns to prioritize salient object boundaries and structural edges while suppressing non-informative regions like floors or walls. 
     }
	\label{fig:scannet_visualization}
    \vspace{-2mm}
\end{figure*}

\vspace{-1mm}
\subsubsection{Autonomous Driving}

\textbf{Baseline.} ReCogDrive~\cite{li2025recogdrive} is a cognitive framework designed for end-to-end autonomous driving.
In our implementation, we focus on its planning capabilities incorporated with \modulelow and VGGT in VLM pre-training stage, without involving the subsequent diffusion planner and reinforcement learning process. 
For evaluation, we conduct experiments on NAVSIM \textit{navtest}~\cite{dauner2024navsim} using closed-loop metrics to assess its driving performance and decision-making intelligence.

\noindent \textbf{Results.} We further evaluate our proposed \modulelow with ReCogDrive on NAVSIM \textit{navtest} using closed-loop metrics. 
As illustrated in \cref{tab:comparison_recogdrive}, \name consistently improves the ReCogDrive baseline across key metrics. Injecting \modulelow during pre-training strengthens spatial awareness and yields significant absolute gains of +3.7\% in DAC (95.5\% vs. 91.8\%) and +2.2\% in TTC (95.0\% vs. 92.8\%). Consequently, these improvements in safety-critical perception lead to a boost in the overall PDMS score, elevating it from 81.6\% to 83.6\%. Overall, the improvements support the effectiveness of \modulelow for enhancing planning-critical spatial reasoning.


\vspace{-1mm}
\subsection{Ablation study}
\label{exp:ablation}
\input{table/06_ablation_study}

We conduct an ablation study on a Qwen2.5-VL-3B backbone to examine the contribution of each component in \name. As shown in \cref{tab:s1_ablation}, the vanilla Qwen2.5-VL-3B baseline achieves 28.66 Avg. Equipping it with \modulelow (SGF) without the \submodulealow (FWC) and \submoduleblow (IG), the model achieves an average score of 47.45, surpassing VG-LLM-4B (46.6) that relies on input-level fusion. This performance gap suggests that input-stage projectors struggle to effectively align fine-grained geometric cues with semantic tokens, whereas SGF preserves geometric information by injecting it directly into the LLM.
Adding \submodulealow (FWC) and \submoduleblow (IG) yields further gains, improving the score to 48.42 and 48.93, respectively. Overall, these gains indicate that enforcing frame-wise constriants and importance-gating helps the model focus geometry integration on task-relevant regions, leading to stronger spatial reasoning.


\vspace{-2mm}
\subsection{Visualization}
\label{exp:visualization}
To better understand how \name utilizes geometric textures, we visualize the importance scores predicted by \submoduleblow, which indicate where the model chooses to emphasize geometry during fusion. As illustrated in \cref{fig:scannet_visualization}, our model naturally learns to prioritize salient objects and structural edges within the scene while significantly down-weighting non-informative regions such as plain floors and walls. 
Notably, this selective focus emerges entirely from training on spatial reasoning tasks without any explicit object mask supervision. This behavior demonstrates that \name interprets spatial environments by identifying key entities and their relational structure, rather than processing the visual field uniformly. This focus concentrates geometry integration on task-relevant structures, consistent with the gains on spatial reasoning benchmarks.

%% file: table/07_video_scaling.tex
\begin{table}[t]
\vspace{-1mm}
\centering
\small
\setlength{\tabcolsep}{0.5mm}
\caption{
\textbf{Performance comparison on general video data mixture.} ($\cdot$) denotes the performance change compared to model trained without general video mixture. Notably, while the pure 2D-based Cambrian-S suffers from performance drops on VSI-Bench due to task interference, \name achieves consistent improvements across both specialized spatial tasks and general video benchmarks with much higher data efficiency.
}
\resizebox{1\linewidth}{!}{
\begin{tabular}{lcccc}
\toprule
\textbf{Model} & \textbf{Video Mixture} & \textbf{VSI} & \textbf{VideoMME} & \textbf{MVBench} \\

\midrule
\multirow{2}{*}{Cambrian-S-7B}
& \xmark & 69.2 & 54.1 & - \\
& 3M & 65.1\textcolor{orange}{\textbf{(-4.1)}} & 61.9\textcolor{blue}{\textbf{(+7.8)}} & 64.5 \\

\midrule
\multirow{2}{*}{\name\textsubscript{Qwen3vl-8B}}
& \xmark & 72.0 & 53.7 & 42.8 \\
& 430k & 72.6\textcolor{blue}{\textbf{(+0.6)}} & 59.4\textcolor{blue}{\textbf{(+5.7)}} & 69.1\textcolor{blue}{\textbf{(+26.3)}} \\

\bottomrule
\end{tabular}}
\label{tab:comparisons_video_scaling}
\vspace{-6mm}
\end{table}

%% file: table/04_VSI_Debiased.tex
\begin{table}[t]
\vspace{-1mm}
\centering
\small
\setlength{\tabcolsep}{1.5mm}
\caption{
\textbf{Performance comparison and frame ablation on VSI and VSI-Debiased.} While Cambrian-S-7B~\cite{yang2025cambrian} is trained on 64/128 frames and VG-LLM-8B*\cite{zheng2025learning(vgllm)} is trained on 8 frames with S1+S2 setting, \name is trained on a maximum of 8/32 frames. We evaluate the zero-shot extrapolation capability of all models by scaling inference frames to 128.
}
\resizebox{1\linewidth}{!}{
\begin{tabular}{llcccc}
\toprule
\multirow{2}{*}{\textbf{Model}} & \multirow{2}{*}{\textbf{Benchmark}} & \multicolumn{4}{c}{\textbf{\# Frames}} \\
\cmidrule(lr){3-6}
 &  & \textbf{16} & \textbf{32} & \textbf{64} & \textbf{128} \\

\midrule
\multirow{2}{*}{Cambrian-S-7B}
& VSI & 58.6 & 63.6 & 66.4 & \sota{67.5} \\
& VSI-Debiased & 49.7 & 55.6 & 59.1 & \sota{59.9} \\
 
\midrule
\multirow{2}{*}{VG-LLM-8B*}
& VSI & 60.5 & 62.2 & \sota{63.7} & 63.1 \\
& VSI-Debiased & 51.6 & 52.4 & \sota{55.2} & 55.1 \\

\midrule
\multirow{2}{*}{\name\textsubscript{Qwen3vl-8B-8frame}}
& VSI & 67.1 & 69.8 & 70.3 & \sota{71.2} \\
& VSI-Debiased & 60.7 & 64.8 & 64.3 & \sota{65.3} \\

\midrule
\multirow{2}{*}{\name\textsubscript{Qwen3vl-8B-32frame}}
& VSI & 69.2 & 72.6 & 73.4 & \sota{73.4} \\
& VSI-Debiased & 64.3 & 66.3 & 67.7 & \sota{68.1} \\

\bottomrule
\end{tabular}}
\vspace{-2pt}
\label{tab:comparisons_vsidebiased}
\vspace{-5mm}
\end{table}

%% file: table/03_RefSpatial.tex
\begin{table*}[ht]
\belowrulesep=0pt
\aboverulesep=0pt
\caption{\textbf{Performance comparisons on RefSpatial-Bench} including the splits of location, placement, and unseen compositional spatial relation. The \sota{bold} and \underline{underlines} values represent the top-1 and top-2 accuracies, respectively. 2B-SFT refers to SFT with a 2B backbone.
}
\centering
\renewcommand\arraystretch{1.1}
\setlength{\tabcolsep}{6pt}
\resizebox{\linewidth}{!}{
\begin{tabular}{l|ccccccccc}
\toprule
\multirow{3}{*}{RefSpatial-Bench}
&  \textit{Proprietary Models}
& \multicolumn{4}{c}{\textit{Referring Specialist Models}}   
& \multicolumn{1}{c}{RoboRefer} 
& \multicolumn{1}{c}{ \cellcolor{navyblue!10} \name \textit{(Ours)}} \\

\cmidrule(r){2-2} \cmidrule(lr){3-6} \cmidrule(lr){7-7} \cmidrule(l){8-8}

& Gemini-2.5-Pro
& SpaceLLaVA
& RoboPoint
& Molmo-7B
& Molmo-72B
& 2B-SFT
& \cellcolor{navyblue!10}2B-SFT \\

\midrule

Location  & 46.96 & 5.82 & 22.87 & 21.91 & 45.77  & \underline{47.00} & \cellcolor{navyblue!10}\sota{48.00}\\
Placement  & 24.21 & 4.31 & 9.27 & 12.85 & 14.74 & \underline{46.00} & \cellcolor{navyblue!10}\sota{47.00} \\
Unseen  & 27.14 & 4.02 & 8.40 & 12.23 & 21.24 & \underline{33.77} & \cellcolor{navyblue!10}\sota{37.66} \\

\midrule
Avg. Acc. & 32.77 & 4.71 & 13.51 & 15.66 & 27.25 & \underline{42.56} & \cellcolor{navyblue!10}\sota{44.22} \\

\bottomrule[1pt]
\end{tabular}}
\label{tab:comparison_roborefer}
\vspace{-4mm}
\end{table*}

%% file: table/05_recogdrive.tex
\begin{table}[t]
    \centering
    \small
    \vspace{1mm}
    \caption{\textbf{Performance comparison on NAVSIM \textit{navtest} using closed-loop metrics.} Evaluation with safety-critical metrics shows that \name enhances planning accuracy over the ReCogDrive, including \underline{N}ot-at-fault \underline{C}ollisions (NC), \underline{D}rivable \underline{A}rea \underline{C}ompliance (DAC), \underline{T}ime-\underline{T}o-\underline{C}ollision within bound (TTC), \underline{Comf}ort (Comf.), \underline{E}go \underline{P}rogress (EP), and \underline{P}edestrian \underline{D}istance \underline{M}argin \underline{S}afety (PDMS).
    }
    \setlength{\tabcolsep}{1.2pt}
    \begin{tabular}{@{}l|cc|ccc|cc@{}}
        \toprule
        Method & NC$\uparrow$ & DAC$\uparrow$ & TTC$\uparrow$ & Comf.$\uparrow$ & EP$\uparrow$ & PDMS$\uparrow$ \\
        \midrule
        Constant Velocity & 68.0 & 57.8 & 50.0 & 100 & 19.4 &  20.6 \\
        Ego Status MLP & 93.0 & 77.3 & 83.6 & 100 & 62.8 &  65.6 \\        
        \midrule
        ReCogDrive w/ InternVL &  97.5 & 91.8 & 92.8 & 100 & 75.0  &  81.6 \\
        \rowcolor{navyblue!10} 
        \name \textit{(Ours)}&  97.0 & 95.5 & 95.0 & 100 & 74.3  & \sota{83.6} \\
        \bottomrule
    \end{tabular}
    \label{tab:comparison_recogdrive}
    \vspace{-5mm}
\end{table}

%% file: table/06_ablation_study.tex
\begin{table}[t] 
    \captionsetup{type=table}
    \vspace{-2mm}
    \centering 
    \caption{\textbf{Ablation Study of Components on VSI-Bench.} \textbf{SGF} denotes our \modulelow, \textbf{CA} denotes the cross-attention with geometric feature, \textbf{FWC} denotes the \submodulealow and \textbf{IG} denotes the \submoduleblow, respectively. }
    \begin{minipage}{0.48\textwidth} 
    \centering 
    \resizebox{\linewidth}{!}{%
        \fontsize{4.6pt}{4.4pt}\selectfont
        \setlength\tabcolsep{3pt}%
        \renewcommand{\arraystretch}{1.4}%
        \begin{tabular}{ccc|c|cccccccc}

        \multicolumn{3}{c|}{\textbf{SGF}} & & 
        \rotatebox{75}{Obj. Count} &
        \rotatebox{75}{Abs. Dist.} &
        \rotatebox{75}{Obj. Size} &
        \rotatebox{75}{Room Size} &
        \rotatebox{75}{Rel. Dist.} &
        \rotatebox{75}{Rel. Dir.} &
        \rotatebox{75}{Route Plan} &
        \rotatebox{75}{Appr. Order} \\

        \cline{1-3}
        
        \textbf{CA} & \textbf{FWC} & \textbf{IG} & Avg. &
        \multicolumn{4}{c}{\cellcolor{orange!10}Numerical Answer} &
        \multicolumn{4}{c}{\cellcolor{yellow!10}Multiple-Choice Answer} \\
        \hline

        \xmarkg & \xmarkg & \xmarkg & 28.66 & 32.7 & 19.5 & 17.3 & 25.1 & 37.3 & 44.9 & 30.4 & 21.8 \\
        
        \cmark & \xmarkg & \xmarkg & 47.45 & 66.5 & 35.8 & 56.5 & 60.0 & 44.3 & 46.9 & 32.9 & 36.4 \\
        
        \cmarkg & \cmark & \xmarkg & 48.42 & 67.5 & 35.3 & 57.7 & 59.6 & 46.0 & 46.9 & 32.9 & 41.1 \\

        
        \rowcolor{navyblue!10} 
        \cmarkg & \cmarkg & \cmark & \sota{48.93} & 68.4 & 36.1 & 57.3 & 62.4 & 43.6 & 47.9 & 34.5 & 40.9 \\
        \hline
        \end{tabular}%
        }  
    \end{minipage}
    \label{tab:s1_ablation}
    \vspace{-5mm}
\end{table}

%% file: section/05_conclusion.tex
\vspace{-2mm}
\section{Conclusion}
\vspace{-1mm}

We presented \name, an active geometry integration framework for enhancing spatial reasoning in MLLMs. Motivated by the limitations of passive fusion, where geometry is treated as a uniformly exposed stream that can induce semantic–geometry misalignment and redundant noise, \name shifts geometry integration from passive fusion to active perception.
Concretely, our Spatial-Grounded Fusion enables semantic visual priors to query task-relevant geometric cues via frame-strict cross-attention, while Importance Gating further concentrates integration on task-relevant regions. Experiments show that \name achieves consistent gains across spatial intelligence benchmarks, setting a new state-of-the-art on VSI-Bench and remaining robust under debiased and long-video evaluation. \name also transfers to downstream tasks, improving RoboRefer and ReCogDrive. These results highlight active geometry integration as a promising path toward spatial intelligence.

\textbf{Acknowledgment.} This work is supported by National Key Research and Development Program of China(2024YFE0203100), Scientific Research Innovation Capability Support Project for Young Faculty (No.ZYGXQNJSKYCXNLZCXM-I28), National Natural Science Foundation of China (NSFC) under Grants No.62476293 and No.62372482,  and General Embodied AI Center of Sun Yat-sen University.

\section{Impact Statement}
This paper presents work whose goal is to advance the field
of Machine Learning. There are many potential societal
consequences of our work, none of which we feel must be
specifically highlighted here.

%% file: section/appendix.tex
\appendix
\clearpage
\newpage

\section{Appendix/supplemental material}
The outline of the Appendix is as follows:
\begin{itemize}
    \item More implementation details;
    \item More analysis on computational cost;
    \item More analysis on fusion ratio $\rho$;
    \item More comparisons on EASI leaderboard; 
    \item More comparisons on VSI-Debiased;
    \item More comparisons on VSTI-Bench; 
    \item More comparisons on GameBench;
    \item More comparisons on general benchmark;
    \item More comparisons on DSR;
    \item More visualization of importance scores;
        \begin{itemize}
            \item Additional visualization on MindCube;
            \item Additional visualization on VSI-Bench;
            \item Robustness to image resolution;
        \end{itemize}
    \item More discussion;
        \begin{itemize}
            \item Additional discussion of scalability to long-range video frames;
            \item Additional discussion of limitation;
            \item Additional discussion of LLM useage;
        \end{itemize}
        
\end{itemize}

\section{Implementation Details}

\subsection{Model Configurations}
We evaluate our method under two primary settings with same setup of learning rate and batch size:
\begin{itemize}
    
    \item \name\textsubscript{Qwen3VL-8B-8frame}: The model is trained with 8 uniformly sampled frames for each scene. Compared to the VG-LLM baseline, the architectural modification is restricted to the inclusion of our Spatial-Grounded Fusion module.

    \item \name\textsubscript{Qwen3VL-8B-32frame}: To handle 32 frames per scene while remaining efficient, we integrate a spatial compression strategy into the SGF framework. While the compressor itself is architecture-agnostic, it functions as a synergetic component to our Importance Gating (IG). By leveraging IG to filter redundant tokens, the framework can employ a larger spatial merge size from 2 to 4 without losing key semantic information that standard architectures would struggle to achieve. We also apply a heuristic bypass for short sequences ($\le$8 frames) to safeguard fine-grained features.

\end{itemize}

\subsection{Data Curation}

For the In-Domain training of our model, we curated a large-scale multimodal dataset totaling 1.8M samples. The data composition is as follows:

\textbf{Spatial Reasoning:} Cambrian-S VSI-bench instruction (590k), SPAR (234k), VLM-3R VSI-bench instruction (205k), VLM-3R VSTI-bench instruction (132k), and MindCube training set (10k).

\textbf{General Video:} LLaVA-Hound (64k), PhysGame PhysInstruct (140k) and a subset of general video data sampled from Cambrian-S-3M (430k).

\subsection{Fusion Ratio $\rho$ and Layer Selection}

The fusion ratio $\rho$, representing the proportion of LLM layers integrated with SGF, is optimized based on the evaluation setting:

\textbf{Out-of-Domain}: We set $\rho$=0.5. To balance semantic reasoning with spatial groundedness, we apply SGF to the middle 50\% of the LLM layers (i.e., range [0.25,0.75]), effectively skipping the initial and final 25\% of layers.

\textbf{In-Domain}: We set $\rho$=0.75 to maximize performance, while consistently skip the final 25\% of LLM layers.

\subsection{Importance Gate Parameter $\epsilon$}

The hyperparameter $\epsilon$ in the Importance Gate modulates the intensity of spatial feature injection. We use $\epsilon$=1e-6 for Out-of-Domain and $\epsilon$=0.1 for In-Domain scenarios.

A smaller $\epsilon$ enforces a stronger, more concentrated control over spatial texture features, which is beneficial for specialized spatial tasks. In contrast, for In-Domain training where general video data is mixed in, a larger $\epsilon$=0.1 is adopted to achieve a smoother control signal, facilitating better generalization across diverse video domains.


\section{Additional analysis of computational cost}

\begin{figure*}[h]
     \centering

     \begin{subfigure}[b]{0.48\textwidth}
         \centering
         \includegraphics[width=\textwidth]{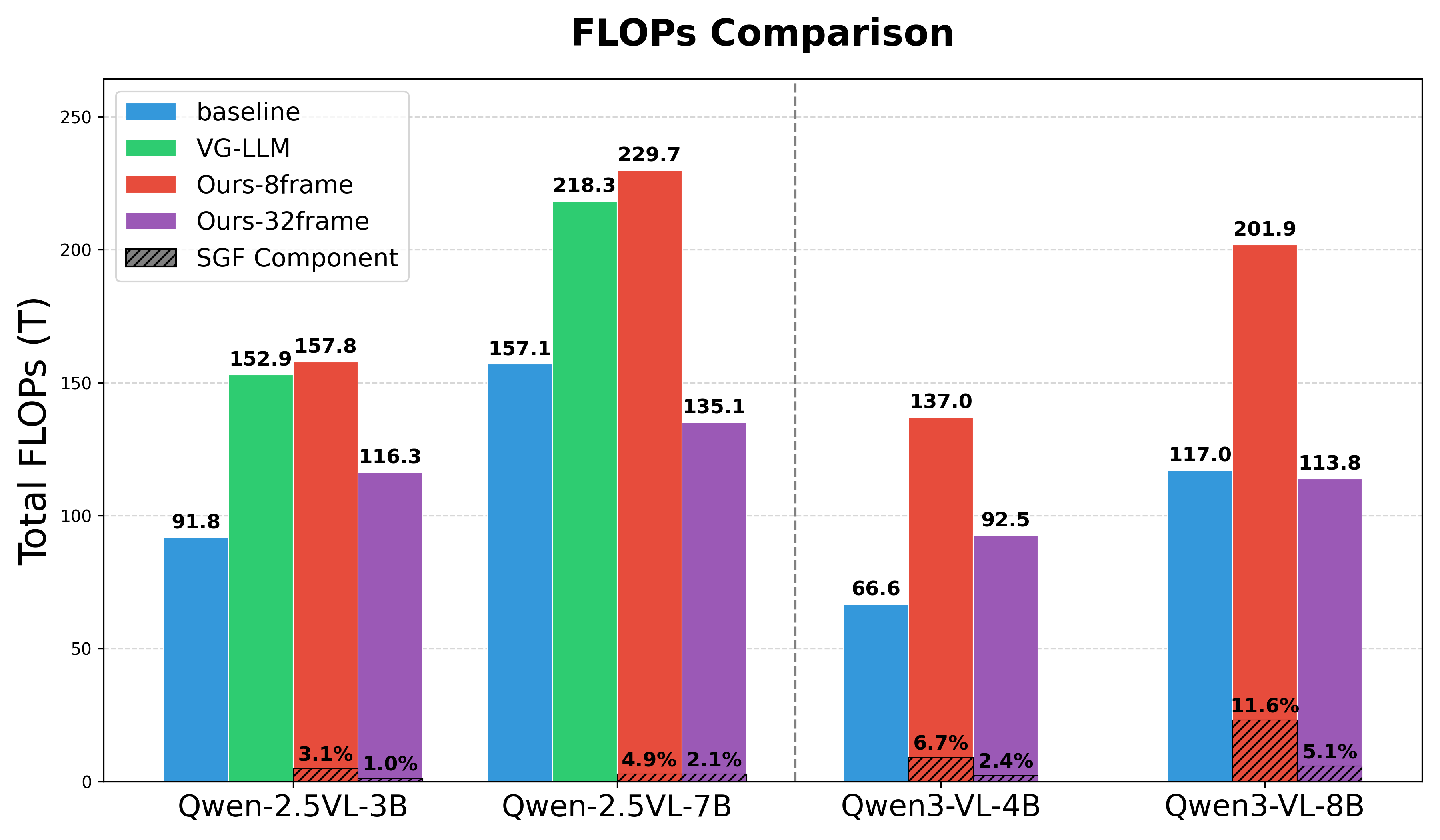}
         \caption{Total FLOPs}
         \label{fig:appendix_compute_cost_a}
     \end{subfigure}
     \hfill 
     \begin{subfigure}[b]{0.48\textwidth}
         \centering
         \includegraphics[width=\textwidth]{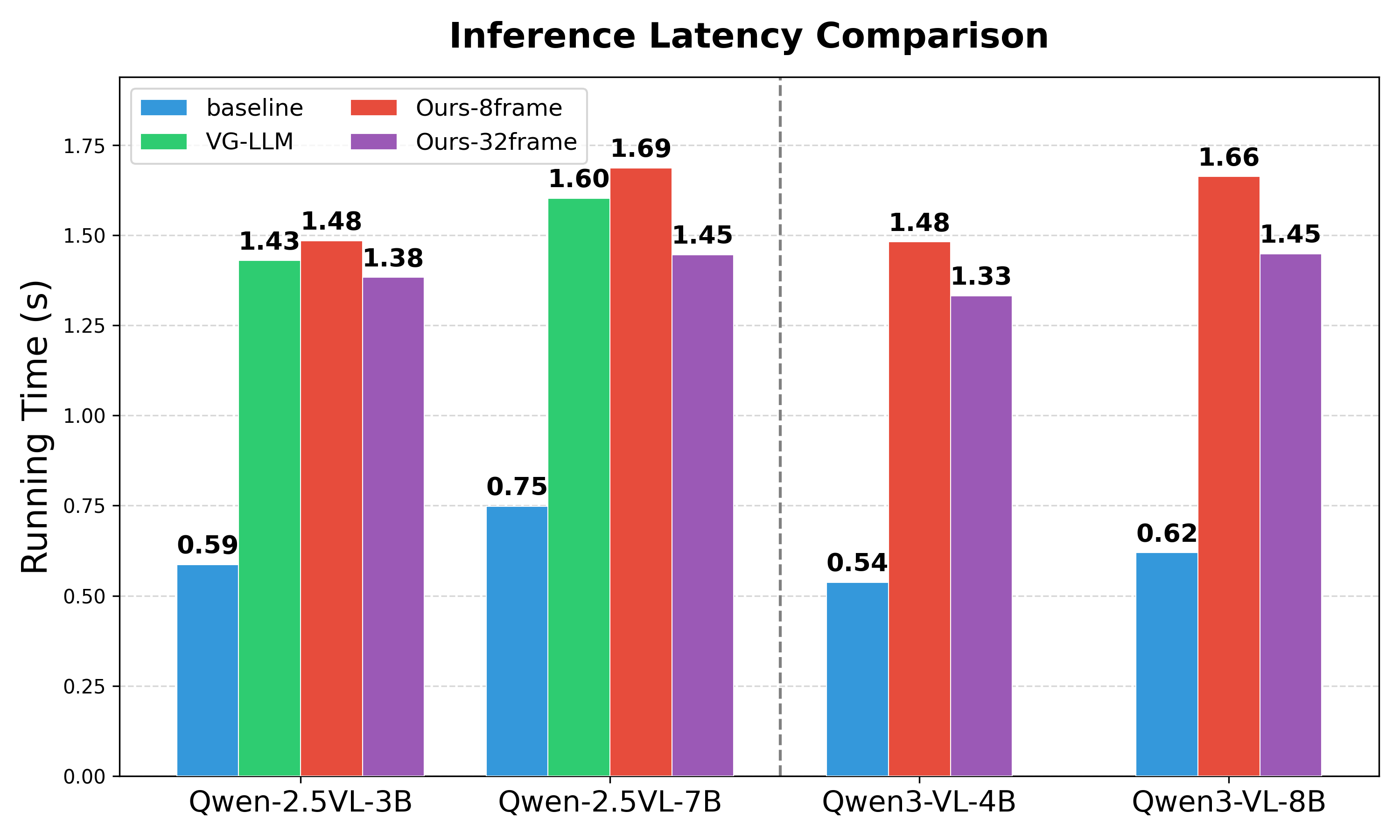}
         \caption{Inference latency}
         \label{fig:appendix_compute_cost_b}
     \end{subfigure}
     
     \caption{Computational cost comparison of FLOPs and inference latency.}
     \label{fig:appendix_compute_cost}
\end{figure*}

To provide a comprehensive evaluation of our model’s efficiency, we analyze the computational cost in terms of Total FLOPs and Inference Latency. We compare our method against the native QwenVL series serving as the baseline and VG-LLM. The evaluation is conducted on the VSI-bench test set and 32 frames are uniformly sampled for each scene.

\subsection{Analysis of Total FLOPs}
As illustrated in \cref{fig:appendix_compute_cost_a}, our proposed SGF module introduces minimal computational overhead:

\textbf{Minimal Overhead of SGF:} On the Qwen-2.5VL backbone series, the FLOPs difference between our 8-frame model and VG-LLM is negligible, with the SGF module accounting for less than 5\% of the total FLOPs. While this proportion slightly increases on the Qwen3-VL series due to differences in hidden state dimensions, the overall efficiency remains high.

\textbf{Efficiency of Spatial Compression:} Our 32-frame setting significantly reduces the total FLOPs through spatial merging. Notably, on larger backbones like Qwen-2.5VL-7B and Qwen3-VL-8B, the Ours-32frame model even achieves lower FLOPs than the original baseline.

\textbf{Conclusion:} These results confirm that the number of visual tokens is the dominant factor influencing total FLOPs, rather than the fusion architecture itself.

\subsection{Analysis of Inference Latency}

\cref{fig:appendix_compute_cost_b} presents the actual running time, revealing the following insights:

\textbf{Comparison with VG-LLM:} In the 8-frame setting, our model exhibits latency nearly identical to VG-LLM, suggesting that the SGF module does not create a bottleneck in the inference pipeline. In the 32-frame setting, our model consistently outperforms VG-LLM in speed due to the effective spatial compression.

\textbf{Sequential Bottleneck:} All models incorporating VGGT are significantly slower than the baseline QwenVL backbone. This is primarily because the 2D image encoder and the VGGT module operate sequentially rather than in parallel. The time consumption is dominated by the VGGT's processing of image features before they enter the LLM.

\textbf{Conclusion:} While our method introduces additional components for spatial intelligence, the use of spatial compression in the 32-frame version provides a superior trade-off between temporal context window and inference speed, making it more practical for long-video understanding than traditional dense sampling methods.


\section{Additional ablation study of fusion ratio $\rho$}

\input{table/08_ablation_rho}

\subsection{Performance Analysis}

As shown in the \cref{tab:appendix_rho_ablation} for the Qwen2.5VL-3B backbone, setting $\rho$=1.0 which integrates SGF into every LLM layer, leads to a catastrophic performance drop, with the average score falling to nearly zero of 0.41. Moderate fusion ratios $\in[0.25,0.5,0.75]$ all yield significant improvements over the baseline.

\textbf{Results with Qwen2.5VL-3B-8frame}: The performance peaks at $\rho$=0.5 (48.93). While $\rho$=0.25 and $\rho$=0.75 are also effective, $\rho$=0.5 provides the best balance between spatial groundedness and linguistic integrity.

\textbf{Results with Qwen2.5VL-7B-frame}: We further validated this on the larger 7B backbone. Consistent with the 3B model, $\rho$=0.5 achieves the highest average score at 50.50, significantly outperforming $\rho$=0.25 at 49.21.

\subsection{Cross-Backbone Insights}

The comparison between the 3B and 7B backbones provides key insights into how model scale affects fusion:

\textbf{On the sensitivity of semantic-geometry fusion:} We observed a performance collapse when integrating SGF into 100\% of the LLM layers ($\rho$=1.0). Qualitative analysis reveals that late-stage integration significantly interferes with the LLM's head-logits, specifically disrupting the prediction of the \text{[EOS]} token. We hypothesize that while intermediate layers are robust enough to internalize task-relevant geometric textures, the final decoding layers are highly specialized for linguistic structure. Injecting external geometric signals at this stage introduces a semantic distribution shift that outweighs the benefits of structural grounding. This discovery validates our Strategic Layer Selection as a crucial mechanism for preserving the generative integrity of MLLMs while enhancing spatial intelligence.

\textbf{Layer Sensitivity:} The Qwen2.5VL-3B model, being smaller in capacity, requires relatively fewer layers to capture the necessary spatial and texture information.

\textbf{Total Layer Depth:} In the Qwen2.5 architecture, the 3B version actually contains more LLM layers (36 layers) compared to the 7B version (28 layers). Consequently, a low ratio like $\rho$=0.25 on the 7B model covers fewer absolute layers than on the 3B model, which may be insufficient to propagate spatial groundedness throughout the network.

\textbf{Conclusion.} Our results demonstrate that a fusion ratio of $\rho$=0.5 is the optimal configuration across different model scales. It provides enough depth for the model to internalize complex spatial-physics relationships without compromising the fundamental instruction-following and termination capabilities of the base LLM.

\input{table/10_EASI_leaderboard}
\section{Additional comparisons on EASI leaderboard}

We evaluate \name\textsubscript{Qwen3-VL-8B-32frame} on the EASI Leaderboard, a comprehensive benchmark for multimodal intelligence. As shown in the \cref{tab:easi_leaderboard}, our model achieves a highly competitive performance, ranking 6-th overall with an average score of 55.0.

\subsection{Data Efficiency}

One of the most significant advantages of \name is its remarkable data efficiency.

\textbf{Comparison with Large-scale Training:} Our model outperforms SenseNova-SI-1.1-QwenVL3-8B (Rank 10) by 2.8 points (55.0 vs. 52.2). Notably, \name\textsubscript{Qwen3-VL-8B-32frame} achieves this superior performance using only 1.8M training samples, whereas the SenseNova variant was trained on a much larger dataset of 8M samples.

\textbf{Insight:} This gap demonstrates that our Spatial-Grounded Fusion architecture and training strategy can extract more effective spatial representations from limited data compared to traditional large-scale pre-training approaches.

\subsection{Substantial Gain over Base Models}

Compared to the original backbone, Qwen3-VL-8B-Instruct (Rank 11), \name provides a substantial performance boost of +7.7 points (55.0 vs. 47.3).

This improvement is particularly evident in benchmarks requiring high-level spatial understanding, such as VSI (72.6 vs. 57.9) and MindCube (83.0 vs. 29.4), where \name nearly triples the score of the base model on MindCube.

This confirms that our architectural enhancements specifically target the deficiencies of existing MLLMs in 3D and spatial intelligence.

\subsection{Analysis of Specialized Benchmarks}

While \name\textsubscript{Qwen3-VL-8B-32frame} shows state-of-the-art capabilities in most spatial tasks, the results also provide insights into areas for further enhancement:

\textbf{MMSI, BLINK, and 3DSRBench:} In these specific benchmarks, our model currently shows room for improvement compared to top-tier proprietary models like Gemini 3 Pro.

\textbf{Future Direction:} The performance on these benchmarks suggests that while our model excels at grounded spatial reasoning, integrating more diverse visual perception tasks or further refining 3D structure-from-motion capabilities could be promising directions for future research. This indicates that the current spatial-grounded features can be further complemented by broader visual-logical reasoning modules.


\section{Additional comparisons on VSI-Debiased}

\input{table/13_appendix_VSI_Debiased}

We further compare our \name with SenseNova-SI\textsubscript{InternVL3-8B}, which is trained with 16 samples per scene. As shown in \cref{tab:appendix_comparisons_vsidebiased}, \name demonstrates strong extrapolation capabilities beyond the training number of frames. \name shows a clear lead over Cambrian-S-7B and SenseNova-SI\textsubscript{InternVL3-8B} even with fewer frames at inference.

\input{table/11_vstibench}
\section{Additional comparisons on VSTI-Bench}

As illustrated in \cref{tab:appendix_vstibench}, our proposed \name achieves an average score of 67.4, securing the 1st rank among all tested models. Notably, it outperforms the leading proprietary model GPT-4o, by a substantial margin of 29.2 points. Compared to the strongest open-source baseline, VLM-3R-7B (58.8), \name demonstrates a significant improvement of 8.6 points, establishing a new state-of-the-art on the VSTI-Bench.

\textbf{Multiple-Choice Answer:} \name exhibits exceptional proficiency in spatial relationship reasoning. In the Object-Object Relative Position task, \name achieves an accuracy of 93.6\%, which is nearly on par with Human Level at 97.5\% and far surpasses GPT-4o at 58.1\%. Similar trends are observed in camera movement direction and relative distance tasks, suggesting that our model possesses a robust internal representation of 3D spatial geometry.

\textbf{Numerical Answer:} Numerical estimation of Absolute Distance and Camera Displacement remains a significant challenge for general-purpose VLMs. While most models, including the 72B-parameter LLaVA-NeXT-Video, struggle with camera displacement. \name achieves a remarkable 45.8. This performance is nearly double that of GPT-4o at 23.4 and approaches the human performance of 46.8, highlighting the effectiveness of our approach in bridging the gap between qualitative perception and quantitative geometric reasoning.

Despite the impressive gains, a gap still exists between \name at 67.4 and Human Level at 77.0, particularly in absolute distance estimation. This suggests that while \name has made significant strides in spatial reasoning, further research is required to achieve human-like precision in complex 3D metric depth estimation.

\section{Additional comparisons on GameBench}

\input{table/12_gamebench}

\cref{tab:appendix_gamebench} presents the evaluation results on PhysGame, a benchmark specifically designed to assess fine-grained physical understanding. Our model demonstrates superior performance across a wide range of physical dimensions. Even without the additional video mixture, \name achieves an average accuracy of 56.9\%, surpassing the previous open-source SOTA, PhysVLM-SFT at 56.7\%, and outperforming leading proprietary models like GPT-4o-0806 at 56.1\%). Specifically, \name shows remarkable strength in understanding Friction (65.9\%) and Body Gesture (66.6\%), highlighting its robust capability in capturing complex physical dynamics.

We investigate the effect of mixing large-scale general video datasets of 430k samples during training. As observed in prior work such as Cambrian-S, incorporating massive amounts of diverse video data can sometimes lead to a slight performance degradation on specialized benchmarks. We observe a similar phenomenon here: the average accuracy drops slightly from 56.9\% (w/o mixture) to 55.7\% (with mixture). We attribute this relatively minor decline to the moderate scale of the PhysGame training set (140k), which maintains a significant influence on the model’s physical understanding capabilities even when blended with larger general datasets. This suggests that while data diversity is crucial, maintaining a balance with domain-specific physical data is key to preserving specialized performance.

It is worth noting that for the PhysGame training and evaluation, we sampled only 8 frames per video scene. Despite this sparse temporal sampling, \name maintains highly competitive performance across all 12 categories, including high-frequency dynamics like Acceleration and Elasticity.


\section{Additional comparisons on general benchmark}

\input{table/14_appendix_general_benchmark}

To verify the general capabilities of \name, we conducted evaluations on several general benchmarks, including AI2D, MMB, and BLINK. As shown in \cref{tab:appendix_general_benchmark}, \name maintains highly competitive performance compared to the general-purpose model Cambrian-S-7B, with no significant performance degradation observed. These results demonstrate that our model has acquired a robust and generalized spatial understanding instead of merely overfitting to specific spatial-related datasets.


\section{Additional comparisons on DSR}

\input{table/15_appendix_DSR}

Furthermore, we evaluated \name on the DSR to assess its effectiveness in dynamic scene understanding. To ensure a fair comparison, we aligned our experimental setup of  model scale and training data with those used in VG-LLM-8B. As illustrated in \cref{tab:appendix_DSR}, \name achieves superior zero-shot performance, outperforming both VLM-3R and VG-LLM. This highlight’s \name’s zero-shot ability to generalize to complex, dynamic environments without task-specific fine-tuning.


\section{Additional visualization of importance scores}

\subsection{Visualization on MindCube}

\begin{figure*}[ht!]
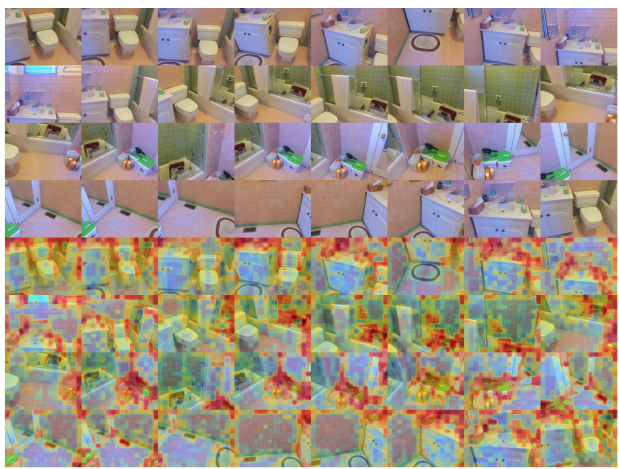

     \centering

     \begin{subfigure}[b]{0.48\textwidth}
         \centering
         \includegraphics[width=\textwidth]{figure/appendix_mindube_visual_1.png}
         \caption{\emph{among\_group458\_q0\_2\_3}}
         \label{fig:appendix_mindube_visual_a}
     \end{subfigure}
     \hfill 
     \begin{subfigure}[b]{0.48\textwidth}
         \centering
         \includegraphics[width=\textwidth]{figure/appendix_mindube_visual_2.png}
         \caption{\emph{among\_group603\_q1\_2\_2}}
         \label{fig:appendix_mindube_visual_b}
     \end{subfigure}
     
     \caption{Visualization of importance score on MindCube.}
     \label{fig:appendix_mindube_visual}
\end{figure*}

\begin{figure*}[ht!]
     \centering

     \begin{subfigure}[b]{0.48\textwidth}
         \centering
         \includegraphics[width=\textwidth]{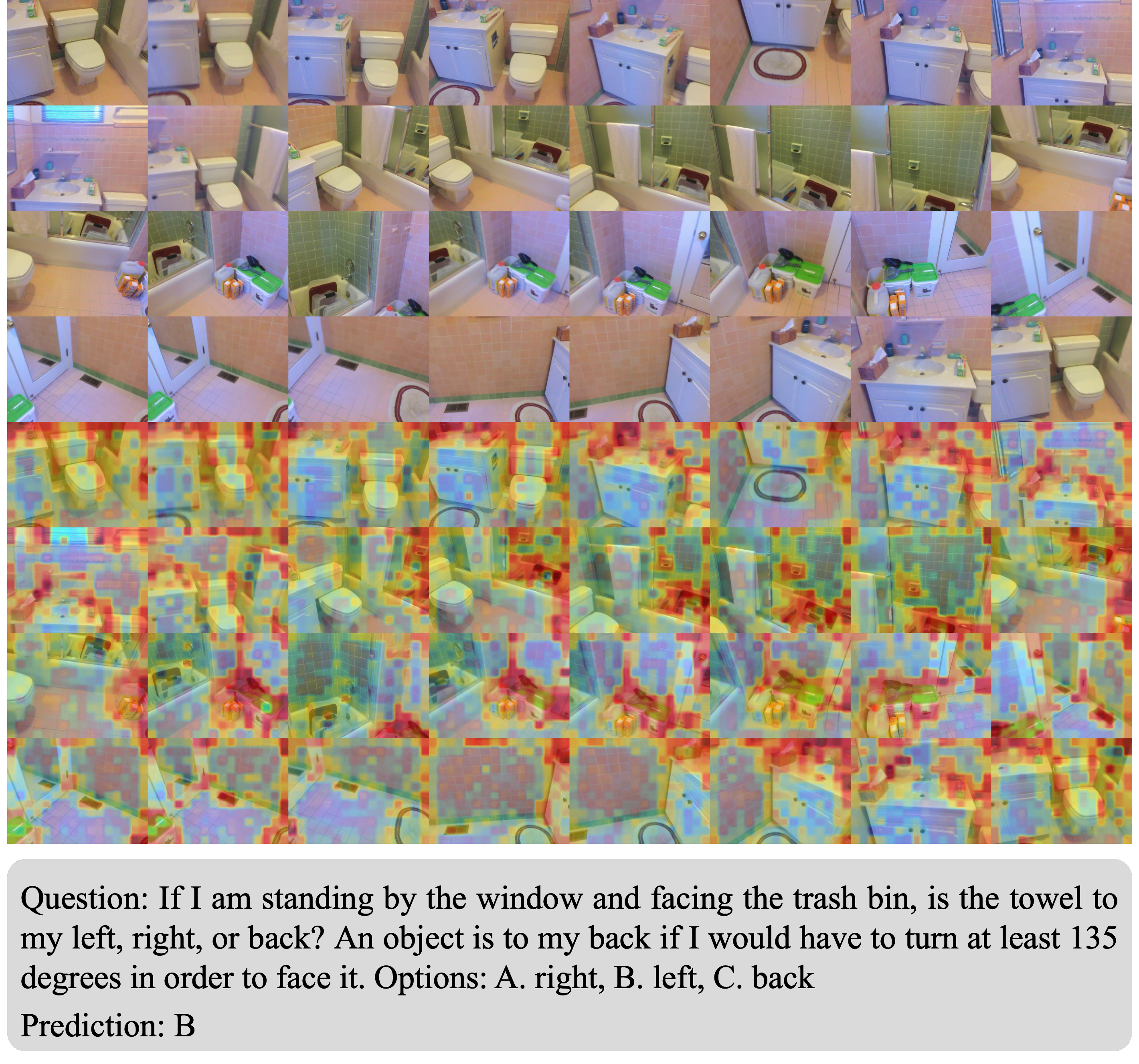}
         \label{fig:appendix_vsibench_visual_a}
     \end{subfigure}
     \hfill 
     \begin{subfigure}[b]{0.48\textwidth}
         \centering
         \includegraphics[width=\textwidth]{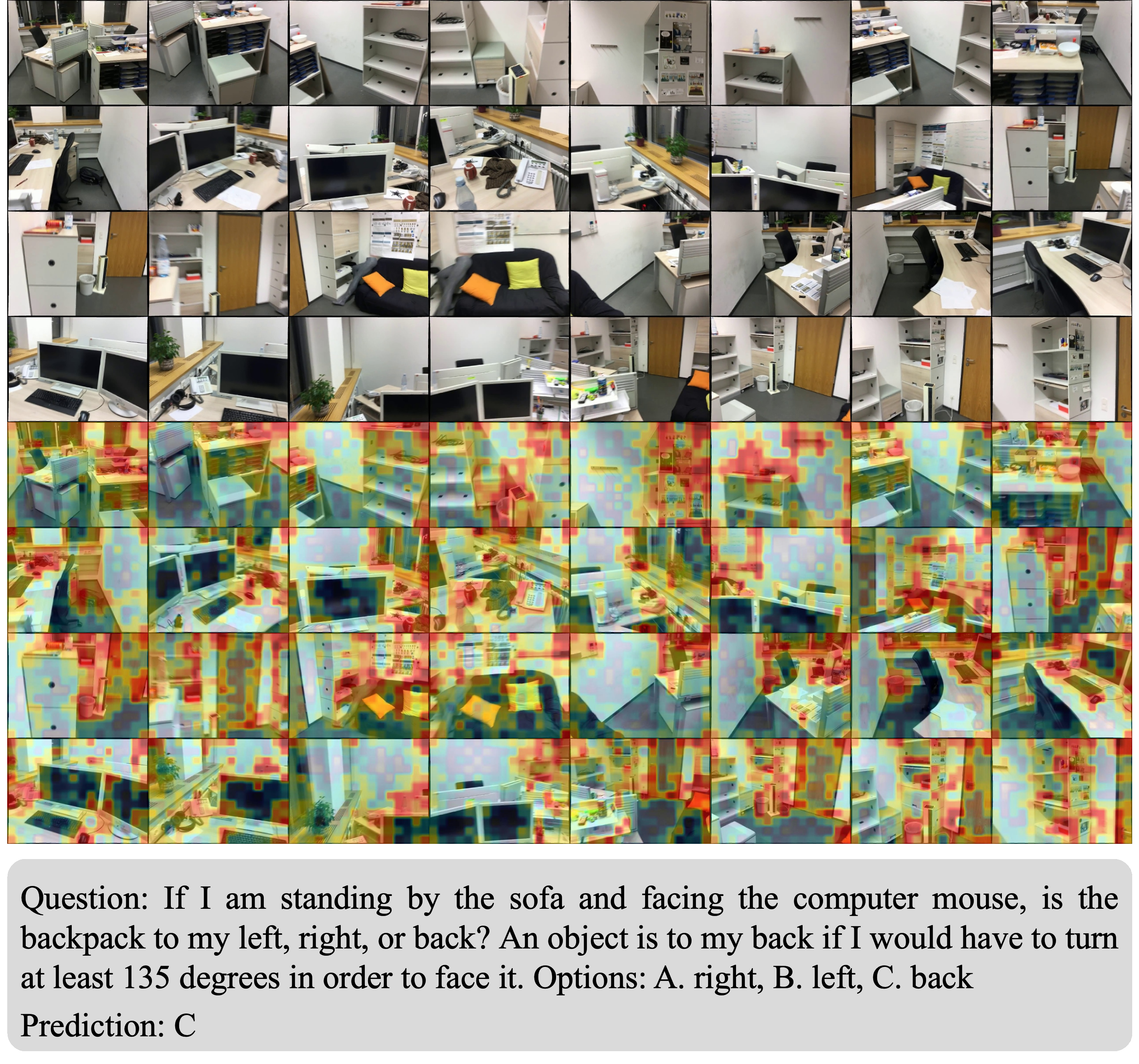}
         \label{fig:appendix_vsibench_visual_b}
     \end{subfigure}
     
     \caption{Visualization of importance score on VSI-Bench.}
     \label{fig:appendix_vsibench_visual}
\end{figure*}

To further investigate the internal reasoning process of \name, we visualize the importance scores (attention maps) on the MindCube benchmark in \cref{fig:appendix_mindube_visual}. MindCube is specifically designed to evaluate a model's spatial intelligence in limited-view scenarios, where the agent must perform complex spatial reasoning based on a set of discrete, non-overlapping viewpoints (front, left, back, and right).

\textbf{Cross-View Information Integration.} As illustrated in \cref{fig:appendix_mindube_visual_a} and \cref{fig:appendix_mindube_visual_b}, when presented with egocentric questions involving multi-step movements ("turn right and move forward"), \name does not merely attend to global image features. Instead, the importance scores are highly concentrated on key semantic landmarks and their surrounding spatial contexts, such as the pink plush toy in the first example and the electric fan in the second. This targeted attention demonstrates the model's ability to \emph{stitch} together a coherent 3D representation from fragmentary 2D views.

\textbf{Grounding Spatial Logic.} The visualization confirms that the model’s correct predictions are grounded in a precise understanding of object-to-object and camera-to-object spatial relationships. Even with a limited field of view, the model successfully identifies the relevant visual cues across different frames to resolve the spatial query.

\subsection{Visualization on VSI-Bench}

\textbf{Fine-grained Object Localization and Grounding.} To qualitatively evaluate our model’s ability to handle dense visual information, we visualize the importance scores on the VSI-bench in \cref{fig:appendix_vsibench_visual}. Unlike the discrete and limited-view nature of MindCube, VSI-bench features highly complex and cluttered indoor environments presented through a continuous stream of frames. As shown in the heatmaps, \name successfully identifies and attends to the specific spatial referents mentioned in the queries, including the towel and trash bin in the bathroom scene, and the backpack and computer mouse in the office setting.

\textbf{Spatial Reasoning via Landmark Identification.} The visualization demonstrates that the model’s spatial reasoning is grounded in precise object localization. In the office example, where the backpack is partially obscured or located among numerous similar desk items, the importance scores are sharply concentrated on the relevant landmarks. This indicates that \name can effectively filter out task-irrelevant visual noise in complex scenes to resolve relative positioning. This ability to pinpoint small, critical objects across multiple frames allows the model to maintain a consistent spatial understanding, even when the viewpoints change rapidly or the environment becomes visually dense.

\begin{figure*}[ht!]
     \centering

     \begin{subfigure}[b]{0.98\textwidth}
         \centering
         \includegraphics[width=\textwidth]{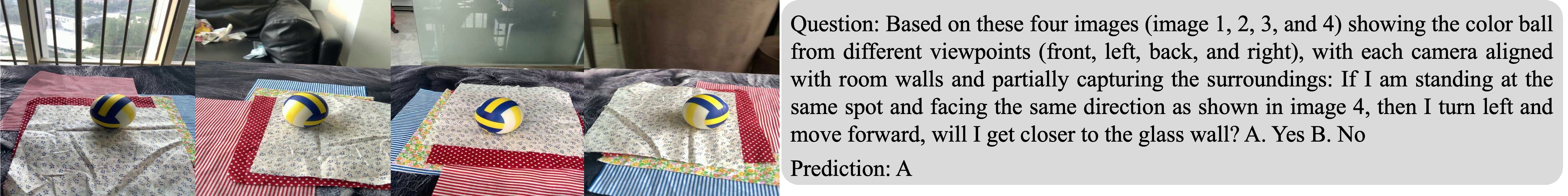}
         \caption{Input question and images with resolution of \text{[448,488]}}
         \label{fig:appendix_resolution_visual_a}
     \end{subfigure}
     \begin{subfigure}[b]{0.98\textwidth}
         \centering
         \includegraphics[width=\textwidth]{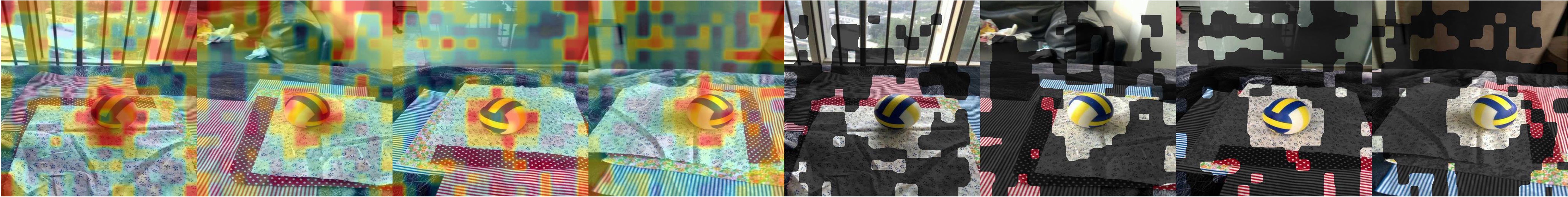}
         \caption{Images with 100\% original resolution}
         \label{fig:appendix_resolution_visual_b}
     \end{subfigure}
     \begin{subfigure}[b]{0.98\textwidth}
         \centering
         \includegraphics[width=\textwidth]{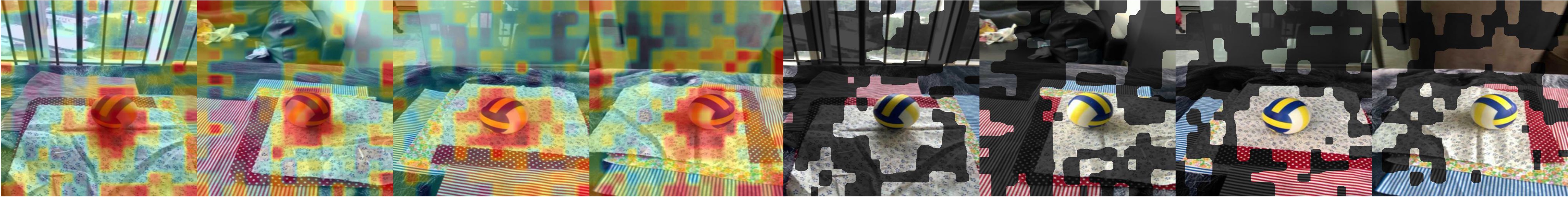}
         \caption{Images with 50\% original resolution}
         \label{fig:appendix_resolution_visual_c}
     \end{subfigure}
     \begin{subfigure}[b]{0.98\textwidth}
         \centering
         \includegraphics[width=\textwidth]{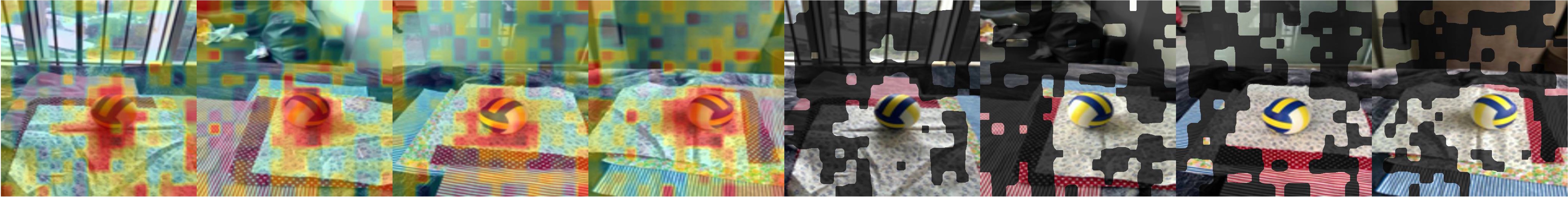}
         \caption{Images with 25\% original resolution}
         \label{fig:appendix_resolution_visual_d}
     \end{subfigure}
     \begin{subfigure}[b]{0.98\textwidth}
         \centering
         \includegraphics[width=\textwidth]{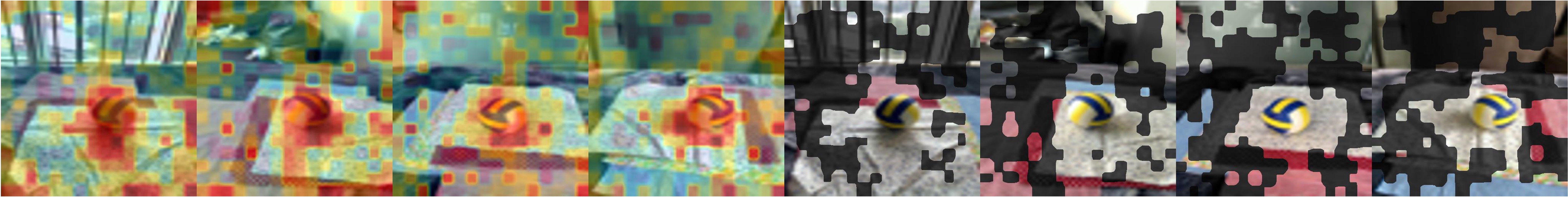}
         \caption{Images with 12.5\% original resolution}
         \label{fig:appendix_resolution_visual_e}
     \end{subfigure}
     \begin{subfigure}[b]{0.98\textwidth}
         \centering
         \includegraphics[width=\textwidth]{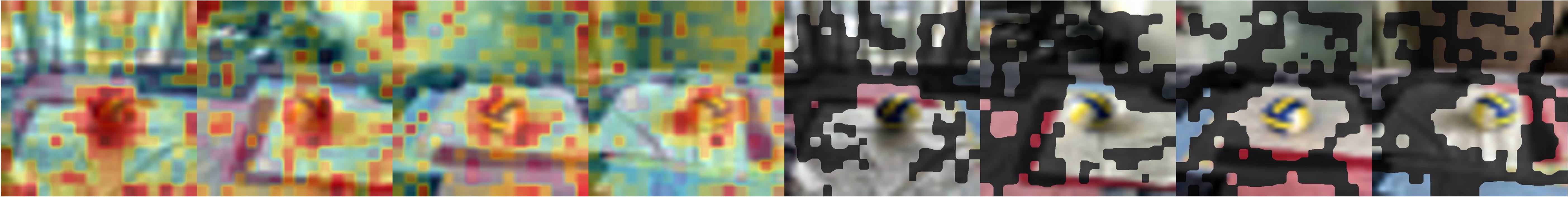}
         \caption{Images with 6.25\% original resolution}
     \end{subfigure}
     
     \caption{\textbf{Visualization of robustness to image resolution.} The left panels show the importance score heatmaps, while the right panels provide a masked visualization where only regions with a heatmap value greater than 0.5 are preserved. The experiment evaluates model performance across varying input quality, from original resolution down to 6.25\%.}
     \label{fig:appendix_resolution_visual}
\end{figure*}

\subsection{Robustness to image resolution}

To evaluate the robustness of our approach, we conducted a downsampling experiment where images were first reduced in resolution and then upsampled back to the original dimensions. This process intentionally discards fine-grained information while maintaining a consistent token count for fair comparison.

As illustrated in \cref{fig:appendix_resolution_visual}, the model consistently maintains its focus on the central object’s texture and key semantic features across all levels of degradation. Remarkably, even when the resolution is aggressively reduced to 6.25\% of the original, which contains only 28×28 pixels of actual information, the model still accurately identifies and attends to the core object. This ability to prioritize essential visual cues despite significant information loss demonstrates the strong robustness of our method against variations in image resolution and its capacity for high-level spatial reasoning.
This demonstrates that our method does not rely solely on high-frequency details but effectively captures essential semantic information, enhancing its strong robustness to variations in image resolution.


\section{Discussions}

\subsection{Scalability to Long-range Video Frames}
we attribute this superior extrapolation capability to the synergy between VGGT-based spatial grounding and our training with general video mixtures. By selectively integrating VGGT features, GeoThinker provides a consistent spatial anchor for each frame. As a result, additional frames are more naturally interpreted as complementary geometric evidence, rather than as redundant noise or out-of-distribution temporal inputs. This enables the model to scale more effectively to longer video contexts.

\subsection{Limitation}
The primary limitation of \name lies in its sensitivity to the accuracy of initial geometric encodings. Information loss at the encoder level can propagate through the fusion modules. Subsequent efforts will focus on developing robust backbones for extreme environments and refining adaptive strategies to enhance the model's capability of thinking with geometry under varied uncertainty.

\subsection{LLM usage}
We thank the Gemini 2.5-Flash for assistance in editing and polishing the manuscript, including grammar checks, sentence structure refinement, and improving overall clarity. The use of this tool did not introduce any new scientific content or ideas. The authors take full responsibility for all content and claims presented in this work.

%% file: table/08_ablation_rho.tex
\begin{table}[h] 
    \captionsetup{type=table}
    \vspace{-3mm}
    \centering 
    \caption{\textbf{Ablation Study of $\rho$ on VSI-Bench (Out-of-Domain).}}
    \begin{minipage}{0.48\textwidth} 
    \centering 
    \resizebox{\linewidth}{!}{%
        \fontsize{4.6pt}{4.4pt}\selectfont
        \setlength\tabcolsep{3pt}%
        \renewcommand{\arraystretch}{1.2}%
        \begin{tabular}{c|c|cccccccc}
        & & 
        \rotatebox{75}{Obj. Count} &
        \rotatebox{75}{Abs. Dist.} &
        \rotatebox{75}{Obj. Size} &
        \rotatebox{75}{Room Size} &
        \rotatebox{75}{Rel. Dist.} &
        \rotatebox{75}{Rel. Dir.} &
        \rotatebox{75}{Route Plan} &
        \rotatebox{75}{Appr. Order} \\
        
        $\rho$ & Avg. &
        \multicolumn{4}{c}{\cellcolor{orange!10}Numerical Answer} &
        \multicolumn{4}{c}{\cellcolor{yellow!10}Multiple-Choice Answer} \\
        \hline
        \rowcolor{black!10}
        \multicolumn{10}{l}{\name(Qwen2.5VL-3B-8frame)} \\
        
        0 & 46.84 & 67.6 & 34.4 & 56.9 & 59.7 & 40.5 & 44.8 & 33.5 & 37.0 \\
        
        0.25 & 48.92 & 68.3 & 36.3 & 57.0 & 60.4 & 47.3 & 47.1 & 35.5 & 39.1 \\

        \rowcolor{navyblue!10}
        0.50 & 48.93 & 68.4 & 36.1 & 57.3 & 62.4 & 43.6 & 47.9 & 34.5 & 40.9 \\
        
       0.75 & 47.86 & 67.5 & 37.0 & 56.9 & 62.3 & 45.0 & 47.5 & 32.4 & 33.9 \\

       1.0 & 0.41 & 2.54 & 0.73 & 0.0 & 0.0 & 0.0 & 0.0 & 0.0 & 0.0 \\

        \hline
        \rowcolor{black!10}
        \multicolumn{10}{l}{\name(Qwen2.5VL-7B-8frame)} \\

        0.25 & 49.21 & 68.7 & 38.6 & 58.3 & 62.0 & 44.2 & 43.5 & 27.8 & 50.3 \\

        \rowcolor{navyblue!10}
        0.50 & 50.50 & 69.5 & 38.5 & 57.9 & 62.2 & 45.2 & 46.2 & 31.4 & 52.6 \\
       
        \hline
        \end{tabular}%
        }  
    \end{minipage}
    \label{tab:appendix_rho_ablation}
\end{table}

%% file: table/10_EASI_leaderboard.tex
\begin{table*}[ht]
  \caption{\textbf{EASI leaderboard (In-Domain).} \emph{Open data source} denotes whether open source data assessment for reproduction, while \cmark represent yes and -- represent the general foundation models.
  VSI denotes VSI-Bench\cite{yang2025thinking(vsi)}. MMSI denotes MMSI-Bench\cite{yang2025mmsi}. MindCube denotes MindCube-Tiny\cite{yin2025spatial(mindcube)}.Viewspatial\cite{li2025viewspatial},SITE\cite{wang2025site}, BLINK\cite{fu2024blink}. EmbSpatial denotes EmbSpatial-Bench\cite{du2024embspatial}. SPAR denotes SPAR-Bench\cite{zhang2025flatland(spar)}. MMSI-Video denotes MMSI-Video-Bench\cite{lin2025mmsi(mmsi-video-bench)}. OmniSpatial\cite{jia2025omnispatial}.
  }
  \label{tab:easi_leaderboard}
  \begin{center}
    \begin{small}
      \begin{sc}
      \setlength{\tabcolsep}{3pt}
        \resizebox{\textwidth}{!}{%
        \begin{tabular}{l|c|cc|ccccccccccc}
          \toprule

          \multirow{2}{*}{\textbf{Models}}
          & \textbf{Open data}
          & \multirow{2}{*}{\textbf{AVG.}}
          & \multirow{2}{*}{\textbf{RANK}}
          & \multirow{2}{*}{\textbf{VSI}}
          & \multirow{2}{*}{\textbf{MMSI}}
          & \multirow{2}{*}{\textbf{MindCube}}
          & \multirow{2}{*}{\textbf{ViewSpatial}}
          & \multirow{2}{*}{\textbf{SITE}}
          & \multirow{2}{*}{\textbf{BLINK}}
          & \multirow{2}{*}{\textbf{3DSRBench}}
          & \multirow{2}{*}{\textbf{EmbSpatial}}
          & \multirow{2}{*}{\textbf{SPAR}}
          & \multirow{2}{*}{\textbf{MMSI-Video}}
          & \multirow{2}{*}{\textbf{OmniSpatial}}\\

          & \textbf{Source} & & & & & & & & & & & & & \\
          

        \midrule
        Gemini 3 Pro & & 1 & 60.8 & 52.5 & 45.2 & 70.9 & 50.4 & 62.2 & 76.0 & 68.9 & 84.3 & 48.7 & 40.4 & 69.1 \\
        Gemini 2.5 Pro & & 2 & 58.0 & 53.6 & 38.0 & 57.6 & 46.1 & 57.1 & 73.5 & 59.3 & 78.8 & -/-  & -/-  & -/-  \\
        SenseNova-SI-1.3-InternVL3-8B & & 3 & 57.3 & 68.6 & 42.5 & 89.9 & 61.3 & 47.5 & 68.0 & 62.4 & 81.0 & 48.4 & 25.7 & 35.3 \\
        SenseNova-SI-1.2-InternVL3-8B & & 4 & 57.0 & 69.6 & 42.6 & 89.0 & 58.8 & 49.0 & 69.4 & 60.1 & 77.7 & 49.5 & 26.2 & 34.8 \\
        GPT-5 & & 5 & 55.7 & 55.0 & 41.8 & 56.3 & 45.6 & 61.9 & 68.0 & 60.3 & 81.6 & 49.7 & 33.4 & 59.2 \\

        \rowcolor{navyblue!13}
        \name\textsubscript{Qwen3VL-8B} & \cmark & 6 & 55.0 & 72.6 & 30.9 & 83.0 & 45.9 & 55.9 & 53.9 & 51.9 & 78.8 &  68.2 & 23.7 & 40.1 \\
        
        Seed 1.6 & & 7 & 54.2 & 49.9 & 38.3 & 48.8 & 43.9 & 54.6 & 65.9 & 56.9 & 75.4 & -/-  & -/-  & -/-  \\
        SenseNova-SI-1.1-InternVL3-8B & & 8 & 54.0 & 68.8 & 43.3 & 85.7 & 54.7 & 47.7 & 63.9 & 55.5 & 72.0 & 45.8 & 23.8 & 33.0 \\
        Grok4 & & 9 & 53.3 & 47.9 & 37.8 & 63.6 & 43.2 & 47.0 & 56.4 & 54.9 & 75.5 & -/-  & -/-  & -/-  \\
        SenseNova-SI-1.1-QwenVL3-8B & & 10 & 52.2 & 64.8 & 38.1 & 73.8 & 51.2 & 49.6 & 61.9 & 53.2 & 72.5 & 40.8 & 25.5 & 43.0 \\
        Qwen3-VL-8B-Instruct & -- & 11 & 47.3 & 57.9 & 31.1 & 29.4 & 42.2 & 45.8 & 66.7 & 53.9 & 77.7 & 39.6 & 28.4 & 47.0 \\
        VST-7B-SFT & \cmark & 12 & 47.2 & 55.5 & 32.5 & 39.7 & 50.5 & 39.7 & 61.9 & 54.6 & 73.7 & 46.6 & 24.9 & 39.5 \\
        SenseNova-SI-1.1-QwenVL2.5-7B & & 13 & 46.5 & 58.1 & 32.8 & 54.7 & 45.5 & 43.9 & 55.3 & 46.3 & 71.4 & 38.2 & 26.1 & 39.3 \\
        InternVL3\_5-8B & -- & 14 & 46.0 & 56.1 & 29.0 & 40.2 & 40.0 & 43.8 & 58.2 & 49.2 & 75.7 & 38.2 & 28.0 & 47.4 \\
        SenseNova-SI-1.1-BAGEL-7B-MoT & & 15 & 45.5 & 41.5 & 34.5 & 46.8 & 46.9 & 42.0 & 65.4 & 42.4 & 69.0 & 44.7 & 23.8 & 44.0 \\
        vlm-3r-llava-qwen2-lora & \cmark & 16 & 44.2 & 60.7 & 27.9 & 40.0 & 40.5 & 31.3 & 52.3 & 51.5 & 68.2 & 42.4 & 27.8 & 43.3 \\
        VST-3B-SFT & \cmark & 17 & 44.1 & 51.4 & 28.8 & 36.0 & 52.9 & 35.9 & 58.8 & 54.1 & 69.0 & 37.7 & 24.3 & 36.5 \\
        SenseNova-SI-1.1-InternVL3-2B & & 18 & 43.6 & 63.7 & 34.2 & 41.8 & 52.7 & 36.8 & 52.4 & 50.5 & 62.8 & 38.0 & 20.4 & 26.4 \\
        InternVL3-8B & -- & 19 & 43.4 & 42.1 & 28.0 & 41.5 & 38.7 & 41.1 & 53.5 & 44.2 & 76.3 & 35.9 & 30.2 & 45.3 \\
        Cambrian-S-7B & \cmark & 20 & 43.3 & 62.9 & 27.1 & 37.9 & 41.3 & 36.1 & 37.9 & 54.8 & 72.8 & 37.9 & 25.2 & 41.9 \\
        BAGEL-7B-MoT & -- & 21 & 42.8 & 31.4 & 31.0 & 34.7 & 41.3 & 37.0 & 63.6 & 50.2 & 73.1 & 39.1 & 27.8 & 41.7 \\ 
        SenseNova-SI-1.1-QwenVL2.5-3B & & 22 & 41.3 & 54.9 & 30.8 & 52.6 & 43.5 & 37.8 & 45.6 & 45.0 & 55.2 & 30.8 & 25.1 & 32.5 \\
        Qwen3-VL-2B-Instruct & -- & 23 & 41.1 & 50.4 & 28.9 & 34.5 & 37.0 & 35.7 & 53.2 & 47.5 & 70.1 & 33.9 & 26.6 & 34.6 \\
        Cambrian-S-3B & \cmark & 24 & 40.4 & 56.1 & 27.0 & 38.4 & 41.0 & 31.0 & 37.7 & 50.9 & 63.5 & 33.0 & 23.9 & 41.9 \\
        Qwen2.5-VL-7B-Instruct & -- & 25 & 39.9 & 32.3 & 26.8 & 36.0 & 36.9 & 37.6 & 55.9 & 43.5 & 71.8 & 33.8 & 27.1 & 37.4 \\
        ViLaSR & \cmark & 26  & 39.5 & 44.6 & 30.2 & 35.1 & 35.7 & 38.7 & 51.4 & 46.6 & 67.3 & 37.4 & 28.3 & 19.2 \\
        SpaceR-SFT-7B & \cmark & 27  & 39.4 & 41.6 & 27.4 & 38.0 & 35.9 & 34.3 & 49.6 & 40.5 & 66.9 & 34.2 & 24.7 & 41.0 \\
        SpatialLadder-3B & \cmark & 28  & 39.1 & 44.9 & 27.4 & 43.5 & 39.9 & 28.0 & 43.0 & 42.8 & 58.2 & 32.9 & 27.4 & 41.9 \\
        Qwen2.5-VL-3B-Instruct & -- & 29  & 38.2 & 27.0 & 28.6 & 37.6 & 32.0 & 33.1 & 48.7 & 53.9 & 62.3 & 28.3 & 27.7 & 41.1 \\
        InternVL3-2B & -- & 30  & 37.9 & 33.0 & 26.5 & 37.5 & 32.6 & 30.0 & 50.8 & 47.7 & 60.1 & 27.2 & 29.1 & 42.0 \\
        Spatial-MLLM-subset-sft & \cmark  & 31  & 35.8 & 46.3 & 26.1 & 33.5 & 34.7 & 18.0 & 40.5 & 36.2 & 50.0 & 35.3 & -/-  & 38.0 \\
        MindCube-Qwen2.5VL-RawQA-SFT & \cmark & 32  & 20.6 & 17.2 & 1.7 & 51.7 & 24.1 & 6.3 & 35.1 & 2.8 & 37.0 & 20.8 & 5.2 & 24.5  \\
          
          \bottomrule
        \end{tabular}}
      \end{sc}
    \end{small}
  \end{center}
  \vskip -0.1in
\end{table*}

%% file: table/13_appendix_VSI_Debiased.tex
\begin{table}[t]
\vspace{-1mm}
\centering
\small
\setlength{\tabcolsep}{1.5mm}
\caption{
\textbf{Performance comparison and frame ablation on VSI and VSI-Debiased.} While Cambrian-S-7B~\cite{yang2025cambrian} is trained on 64/128 frames and SenseNova-SI\textsubscript{InternVL3-8B}\cite{cai2025scaling(sensenova)} is trained on 16 frames, \name is trained on a maximum of 8/32 frames. We evaluate the zero-shot extrapolation capability of all models by scaling inference frames to 128.
}
\resizebox{1\linewidth}{!}{
\begin{tabular}{llcccc}
\toprule
\multirow{2}{*}{\textbf{Model}} & \multirow{2}{*}{\textbf{Benchmark}} & \multicolumn{4}{c}{\textbf{\# Frames}} \\
\cmidrule(lr){3-6}
 &  & \textbf{16} & \textbf{32} & \textbf{64} & \textbf{128} \\

\midrule
\multirow{2}{*}{Cambrian-S-7B}
& VSI & 58.6 & 63.6 & 66.4 & \sota{67.5} \\
& VSI-Debiased & 49.7 & 55.6 & 59.1 & \sota{59.9} \\

\midrule
\multirow{2}{*}{VG-LLM-8B*}
& VSI & 60.5 & 62.2 & \sota{63.7} & 63.1 \\
& VSI-Debiased & 51.6 & 52.4 & \sota{55.2} & 55.1 \\

\midrule
\multirow{2}{*}{SenseNova-SI\textsubscript{InternVL3-8B}}
& VSI & 64.6 & 68.7 & \sota{68.8} & 66.3 \\
& VSI-Debiased & 58.9 & \sota{62.8} & 62.4 & 59.7 \\

\midrule
\multirow{2}{*}{\name\textsubscript{Qwen3vl-8B-8frame}}
& VSI & 67.1 & 69.8 & 70.3 & \sota{71.2} \\
& VSI-Debiased & 60.7 & 64.8 & 64.3 & \sota{65.3} \\

\midrule
\multirow{2}{*}{\name\textsubscript{Qwen3vl-8B-32frame}}
& VSI & 69.2 & 72.6 & 73.4 & \sota{73.4} \\
& VSI-Debiased & 64.3 & 66.3 & 67.7 & \sota{68.1} \\

\bottomrule
\end{tabular}}
\vspace{-2pt}
\label{tab:appendix_comparisons_vsidebiased}
\vspace{-5mm}
\end{table}

%% file: table/11_vstibench.tex
\definecolor{oai-white}{HTML}{FFFFFF}
\definecolor{oai-black}{HTML}{000000}
\definecolor{oai-red}{HTML}{FF4500}
\definecolor{oai-green}{HTML}{51DA4C}
\definecolor{oai-blue}{HTML}{0000FF}
\definecolor{oai-yellow}{HTML}{FFF639}
\definecolor{oai-magenta}{HTML}{FF45FF}
\definecolor{oai-cyan}{HTML}{00FFFF}
\definecolor{oai-orange}{HTML}{FE7600}
\definecolor{oai-violet}{HTML}{8A2BE2}
\definecolor{oai-brown}{HTML}{A0522D}
\definecolor{oai-green-050}{HTML}{F4FFF4}
\definecolor{oai-green-100}{HTML}{E9FFE8}
\definecolor{oai-green-200}{HTML}{D9FFD8}
\definecolor{oai-green-300}{HTML}{C9FFC7}
\definecolor{oai-green-400}{HTML}{A6FFA3}
\definecolor{oai-green-500}{HTML}{7CF178}
\definecolor{oai-green-600}{HTML}{51DA4C}
\definecolor{oai-green-700}{HTML}{3FA93B}
\definecolor{oai-green-800}{HTML}{2D712A}
\definecolor{oai-green-900}{HTML}{193718}
\definecolor{oai-gray-000}{HTML}{FFFFFF}
\definecolor{oai-gray-100}{HTML}{FAFAFA}
\definecolor{oai-gray-200}{HTML}{F5F5F5}
\definecolor{oai-gray-300}{HTML}{E5E5E5}
\definecolor{oai-gray-400}{HTML}{FFB7A4}
\definecolor{oai-gray-500}{HTML}{CDCDCD}
\definecolor{oai-gray-600}{HTML}{A8A8A8}
\definecolor{oai-gray-700}{HTML}{747474}
\definecolor{oai-gray-800}{HTML}{393939}
\definecolor{oai-gray-900}{HTML}{000000}

\begin{table}[t!]  
    \captionsetup{type=table}
    \centering
    \centering
    \fontsize{4.6pt}{4.4pt}\selectfont 
    \setlength\tabcolsep{3pt}  
    \renewcommand{\arraystretch}{1.4}%
    \caption{
    \textbf{Performance comparison on the VSTI-Bench.} \name\textsubscript{Qwen3VL-8B} achieves the highest average score among all models, significantly outperforming both proprietary and open-source counterparts.The \sota{bold} and \underline{underlines} values represent the top-1 and top-2 accuracies, respectively.
    }
    \scalebox{1.4}{
    \begin{tabular}{r|c|ccccc} 
    & & 
    \rotatebox{75}{Cam-Obj Abs. Dist.} & 
    \rotatebox{75}{Cam. Displace.} &     
    \rotatebox{75}{Cam. Mov. Dir.} &     
    \rotatebox{75}{Obj-Obj Rel. Pos.} &  
    \rotatebox{75}{Cam-Obj Rel. Dist.} \\ 
    Methods & Avg. &
    \multicolumn{2}{c}{\cellcolor{orange!10}Numerical Answer} & 
    \multicolumn{3}{c}{\cellcolor{yellow!10}Multiple-Choice Answer} \\ 
    \hline
    \rowcolor{black!7}
    \multicolumn{7}{l}{\textcolor{black}{\textit{Baseline}}} \\ 
    Chance Level (Random) & - & - & - & 36.1 & 50.0 & 36.1 \\
    Chance Level (Frequency) & 27.4 & 5.4 & 6.2 & 40.7 & 52.2 & 32.4 \\ 
    \hline
    \rowcolor{black!7}
    \multicolumn{7}{l}{\textcolor{black}{\textit{Human Performance}}} \\
    \dag Human Level & 77.0 & 51.4 & 46.8 & 95.1 & 97.5 & 94.3 \\
    \hline
    \rowcolor{black!7}
    \multicolumn{7}{l}{\textcolor{black}{\textit{Proprietary Models (API)}}} \\
    GPT-4o & 38.2 & 29.5 & 23.4 & 37.3 & 58.1 & 42.5 \\ 
    Gemini-1.5 Flash & 32.1 & 28.5 & 20.9 & 24.4 & 52.6 & 33.9 \\
    \hline
    \rowcolor{black!7}
    \multicolumn{7}{l}{\textcolor{black}{\textit{Open-sourced VLMs}}} \\
    LLaVA-OneVision-0.5B& 36.9 & 16.5 & 32.4 & 46.1 & 50.5 & 39.0 \\
    InternVL2-2B & 38.1 & 17.7 & 27.8 & 43.0 & 54.9 & 47.2 \\
    \hline 
    LLaVA-NeXT-Video-7B & 40.0 & 28.2 & 1.8 & 49.8 & 64.7 & 55.6 \\
    LLaVA-OneVision-7B& 41.7 & 29.9 & 19.3 & 47.5 & 62.1 & 49.8 \\    
    LongVA-7B& 32.3 & 13.5 & 5.1 & 43.7 & 57.9 & 41.2 \\
    InternVL2-8B  & 43.5 & 32.9 & 13.5 & 48.0 & 68.0 & 55.0 \\    
    LongVILA-8B  & 30.5 & 20.0 & 11.6 & 35.4 & 52.3 & 33.4 \\
    VILA-1.5-8B & 37.3 & 30.1 & 27.3 & 42.2 & 50.4 & 36.7 \\
    VILA-1.5-40B & 38.2 & 28.2 & 15.7 & 28.8 & 65.4 & 53.0 \\
    LLaVA-NeXT-Video-72B & 44.0 & 32.3 & 10.5 & 48.1 & 78.3 & 50.9 \\    
    VLM-3R-7B & \underline{58.8} & \sota{39.4} & \underline{39.6} & \underline{60.6} & \underline{86.5} & \underline{68.6} \\
    \hline
    \rowcolor{navyblue!10}
    \multicolumn{7}{l}{\textcolor{black}{\textit{Ours}}} \\
    
    \name\textsubscript{Qwen3VL-8B} & \sota{67.4} & \underline{38.4} & \sota{45.8} & \sota{84.2} & \sota{93.6} & \sota{75.2} \\
    \hline
    \end{tabular}
    } 
\label{tab:vsti} 
    \label{tab:appendix_vstibench}
\end{table}

%% file: table/12_gamebench.tex
\begin{table*}[ht]
\centering
\caption{\textbf{Evaluation results (\%) of open-source and proprietary multi-modal LLMs on PhysGame.} The fine-grained categories include gravity, elasticity, friction, velocity, acceleration, reflection, refraction, absorption \& transmission, color, rigidity, object shape, and body gesture. AVG denotes the average accuracy.}
\vspace{-2mm}
\label{tab:appendix_gamebench}
\scalebox{0.8}{
\begin{tabular}{ll|c|cccccccccccc}
\toprule
\multicolumn{1}{c}{\multirow{2}{*}{\textbf{Models}}} & & \multicolumn{1}{c|}{\multirow{2}{*}{\textbf{AVG}}} & \multicolumn{3}{c|}{\textbf{Mechanics}} & \multicolumn{2}{c|}{\textbf{Kinematics}} & \multicolumn{3}{c|}{\textbf{Optics}} & \multicolumn{4}{c}{\textbf{Material}}  \\ 
\cmidrule(r){4-6} \cmidrule(r){7-8} \cmidrule(r){9-11} \cmidrule(r){12-15}
& & & Grav.    & Elast.  & Fric.  & Velo.    & Acc. & Refl.   & Refr. & Abs. & Col. & Rig. & Sha. & Gest.  \\ 
\midrule
\rowcolor{black!8}
\multicolumn{15}{c}{\emph{\textbf{Proprietary Multi-modal LLMs}}}  \\ 
Claude3.5-Sonnet & \cite{anthropic2024claude} & 54.3 & \textbf{50.7} & 58.8 & 50.6 & \textbf{53.2} & 59.1 & \textbf{50.0} & 50.0 & 49.2 & 64.4 & 52.7 & 50.0 & \textbf{62.1} \\ 
Claude3.5-SonnetV2& \cite{anthropic2024claude} & 47.6 & 46.5 & 52.5 & 46.6 & 37.2 & 53.4 & 47.8 & 50.0 & 33.9 & 55.6 & 54.1 & 43.8 & 51.7\\
Gemini-1.5-pro &\cite{team2024gemini(gemini1.5)} & 55.2 & \textbf{50.7} & \textbf{70.0} & 48.9 & 51.1 & 59.1 & \textbf{50.0} & 42.9 & \textbf{52.5} & \textbf{71.1} & \textbf{56.8} & 53.1 & 58.6 \\
Gemini-1.5-pro-flash &\cite{team2024gemini(gemini1.5)} & 48.5 & 47.9 & 52.5 & 51.7 & 43.6 & 51.1 & 43.5 & 53.6 & 33.9 & 64.4 & 43.2 & 46.9 & 49.4 \\
GPT-4V &\cite{achiam2023gpt} & 45.9 & 40.8 & 60.0 & 48.3 & 34.0 & 48.9 & 43.5 & 46.4 & 42.4 & 53.3 & 45.9 & 37.5 & 44.8 \\
GPT-4o-0806 &\cite{hurst2024gpt4o} & 56.1 & 47.9 & 61.3 & \textbf{59.1} & 43.6 & \textbf{61.4} & 43.5 & 53.6 & 50.8 & 68.9 & 54.1 & \textbf{65.6} & 63.2 \\
GPT-4o-mini-0718 &\cite{hurst2024gpt4o} & 40.3 & 43.7 & 43.8 & 39.2 & 35.1 & 44.3 & 30.4 & 46.4 & 42.4 & 44.4 & 37.8 & 37.5 & 41.4 \\
Qwen-VL-max &\cite{bai2023qwen} & 50.9 & \textbf{50.7} & 53.8 & 51.1 & 31.9 & 46.6 & \textbf{50.0} & \textbf{60.7} & 50.8 & 64.4 & 48.6 & \textbf{65.6} & 59.8 \\

\midrule
\rowcolor{black!8}
\multicolumn{15}{c}{\emph{\textbf{Open-source Multi-modal LLMs}}}  \\ 
LLaVA-Next-Video &\cite{liu2024llavanext} & 32.2 & 43.7 & 33.8 & 27.3 & 34.0 & 22.7 & 21.7 & 35.7 & 23.7 & 35.6 & 41.9 & 34.4 & 37.9 \\
Video-LLaVA &\cite{lin2024video} & 29.0 & 32.4 & 22.5 & 27.8 & 31.9 & 26.1 & 19.6 & 35.7 & 32.2 & 31.1 & 36.5 & 28.1 & 27.6 \\
LLaVA-OneVision &\cite{li2024llava-onevision} & 47.7 & 50.7 & 50.0 & 46.0 & 39.4 & 45.5 & 43.5 & \textbf{71.4} & \textbf{40.7} & 55.6 & 44.6 & \textbf{56.2} & 52.9 \\
InternVL2 &\cite{chen2024internvl2} & 33.4 & 29.6 & 31.2 & 38.6 & 35.1 & 30.7 & 30.4 & 53.6 & 35.6 & 26.7 & 29.7 & 18.8 & 34.5 \\
VideoChat2 & \cite{li2024mvbench} & 34.3 & 33.8 & 35.0 & 29.5 & 41.5 & 28.4 & 28.3 & 32.1 & 33.9 & 33.3 & 41.9 & 21.9 & 44.8 \\
ST-LLM & \cite{liu2024st} & 32.8 & 32.4 & 26.2 & 26.7 & 37.2 & 28.4 & 37.0 & 25.0 & 28.8 & 33.3 & 40.5 & 37.5 & 46.0 \\
Chat-UniVi & \cite{jin2024chat} & 29.5 & 28.2 & 27.5 & 29.5 & 39.4 & 23.9 & 28.3 & 32.1 & 30.5 & 31.1 & 18.9 & 28.1 & 35.6 \\
PPLLaVA &  \cite{liu2024ppllava} & 38.4 & 45.1 & 38.8 & 42.6 & 30.9 & 30.7 & 41.3 & 39.3 & 35.6 & 44.4 & 39.2 & 18.8 & 43.7 \\
PhysVLM-SFT & \cite{cao2024physgame}  & 56.7 & 54.9 & 62.5 & \textbf{60.2} & 51.1 & \textbf{63.6} & \textbf{45.7} & 57.1  & 28.8 & \textbf{64.4} & 51.4 & 50.0 & 72.4 \\

\midrule
\rowcolor{navyblue!10}
\multicolumn{15}{c}{\emph{\textbf{Ours}}}  \\ 

\multicolumn{2}{l|}{\name\textsubscript{Qwen3VL-8B} w/o. 430k Video Mixture} & 56.9 & 53.5 & 62.5 & 61.3 & 55.3 & 52.2 & 45.6 & 60.7 & 50.8 & 66.6 & 48.6 & 59.3 & 66.6 \\

\name\textsubscript{Qwen3VL-8B} & & 55.7 & 56.3 & 61.2 & 65.9 & 48.9 & 62.5 & 43.4 & 53.5 & 47.4 & 68.8 & 47.2 & 46.8 & 66.6 \\

\bottomrule
\end{tabular}
}
\end{table*}

%% file: table/14_appendix_general_benchmark.tex
\begin{table}[h]
\centering
\small
\setlength{\tabcolsep}{1.5mm}
\caption{
\textbf{Performance comparison on general benchmark.} 
}
\begin{tabular}{l|ccc}
\toprule
\textbf{Model} & \textbf{AI2D} & \textbf{MMB} & \textbf{BLINK} \\

\midrule

Cambrian-S-7B & 76.9 & 80.4 & 37.9 \\
\name\textsubscript{Qwen3VL-8B-Scaled} & 76.3 & 83.3 & 53.9 \\
 
\bottomrule
\end{tabular}
\label{tab:appendix_general_benchmark}
\vspace{-3mm}
\end{table}

%% file: table/15_appendix_DSR.tex
\begin{table}[h]
\vspace{-3mm}
\centering
\small
\setlength{\tabcolsep}{1.5mm}
\caption{
\textbf{Performance comparison on DSR}\cite{zhou2025learning(DSR)}. 
}
\begin{tabular}{l|ccc}
\toprule
 & \textbf{VLM-3R} & \textbf{VG-LLM-8B} & \textbf{\name}\\
\midrule
Performance & 31.4 & 38.4 & 42.7 \\
 
\bottomrule
\end{tabular}
\label{tab:appendix_DSR}
\vspace{-3mm}
\end{table}

%% file: example_paper.bib
@article{wang2025site,
  title={SITE: towards Spatial Intelligence Thorough Evaluation},
  author={Wang, Wenqi and Tan, Reuben and Zhu, Pengyue and Yang, Jianwei and Yang, Zhengyuan and Wang, Lijuan and Kolobov, Andrey and Gao, Jianfeng and Gong, Boqing},
  journal={arXiv preprint arXiv:2505.05456},
  year={2025}
}

@article{brown2025benchmark,
  title={Benchmark Designers Should" Train on the Test Set" to Expose Exploitable Non-Visual Shortcuts},
  author={Brown, Ellis and Yang, Jihan and Yang, Shusheng and Fergus, Rob and Xie, Saining},
  journal={arXiv preprint arXiv:2511.04655},
  year={2025}
}

@article{jia2025omnispatial,
  title={OmniSpatial: Towards Comprehensive Spatial Reasoning Benchmark for Vision Language Models},
  author={Jia, Mengdi and Qi, Zekun and Zhang, Shaochen and Zhang, Wenyao and Yu, Xinqiang and He, Jiawei and Wang, He and Yi, Li},
  journal={arXiv preprint arXiv:2506.03135},
  year={2025}
}

@article{yang2025mmsi,
  title={MMSI-Bench: A Benchmark for Multi-Image Spatial Intelligence},
  author={Yang, Sihan and Xu, Runsen and Xie, Yiman and Yang, Sizhe and Li, Mo and Lin, Jingli and Zhu, Chenming and Chen, Xiaochen and Duan, Haodong and Yue, Xiangyu and others},
  journal={arXiv preprint arXiv:2505.23764},
  year={2025}
}

@article{li2025viewspatial,
  title={ViewSpatial-Bench: Evaluating Multi-perspective Spatial Localization in Vision-Language Models},
  author={Li, Dingming and Li, Hongxing and Wang, Zixuan and Yan, Yuchen and Zhang, Hang and Chen, Siqi and Hou, Guiyang and Jiang, Shengpei and Zhang, Wenqi and Shen, Yongliang and others},
  journal={arXiv preprint arXiv:2505.21500},
  year={2025}
}

@article{seed2025seed1_5vl,
  title={Seed1.5-VL Technical Report},
  author={{ByteDance Seed}},
  journal={arXiv preprint arXiv:2505.07062},
  year={2025}
}

@article{team2023gemini,
  title={Gemini: a family of highly capable multimodal models},
  author={{Gemini Team}},
  journal={arXiv preprint arXiv:2312.11805},
  year={2023}
}

@misc{gemini_3_pro_systemcard,
  author       = {{Gemini}},
  title        = {{Gemini 3 Pro Model Card}},
  howpublished = {Technical report, Gemini},
  note         = {Accessed: 2025-11-18},
  year         = {2025},
  month        = nov,
  day          = 18,
}

@article{hurst2024gpt4o,
  title={Gpt-4o system card},
  author={Hurst, Aaron and Lerer, Adam and Goucher, Adam P and Perelman, Adam and Ramesh, Aditya and Clark, Aidan and Ostrow, AJ and Welihinda, Akila and Hayes, Alan and Radford, Alec and others},
  journal={arXiv preprint arXiv:2410.21276},
  year={2024}
}

@misc{openai_gpt5_systemcard,
  author       = {{OpenAI}},
  title        = {{GPT-5 System Card}},
  howpublished = {Technical report, OpenAI},
  note         = {Accessed: 2025-08-10},
  year         = {2025},
  month        = aug,
  day          = 7,
}

@article{Qwen2.5-VL,
  title={Qwen2.5-VL Technical Report},
  author={{Qwen Team}},
  journal={arXiv preprint arXiv:2502.13923},
  year={2025}
}

@misc{yang2025qwen3technicalreport,
      title={Qwen3 Technical Report}, 
      author={{Qwen Team}},
      year={2025},
      eprint={2505.09388},
      archivePrefix={arXiv},
      primaryClass={cs.CL},
      url={https://arxiv.org/abs/2505.09388}, 
}

@misc{Qwen3-VL,
  author       = {{Qwen Team}},
  title        = {Qwen3-VL: Multimodal large language model series},
  howpublished = {\url{https://github.com/QwenLM/Qwen3-VL}},
  year         = {2025},
  note         = {GitHub repository; accessed: 2025-11-14}
}

@misc{liu2024llavanext,
    title={LLaVA-NeXT: Improved reasoning, OCR, and world knowledge},
    url={https://llava-vl.github.io/blog/2024-01-30-llava-next/},
    author={Liu, Haotian and Li, Chunyuan and Li, Yuheng and Li, Bo and Zhang, Yuanhan and Shen, Sheng and Lee, Yong Jae},
    month={January},
    year={2024}
}

@article{li2024llava-onevision,
  title={Llava-onevision: Easy visual task transfer},
  author={Li, Bo and Zhang, Yuanhan and Guo, Dong and Zhang, Renrui and Li, Feng and Zhang, Hao and Zhang, Kaichen and Zhang, Peiyuan and Li, Yanwei and Liu, Ziwei and others},
  journal={arXiv preprint arXiv:2408.03326},
  year={2024}
}

@article{chen2024internvl2,
  title={How far are we to gpt-4v? closing the gap to commercial multimodal models with open-source suites},
  author={Chen, Zhe and Wang, Weiyun and Tian, Hao and Ye, Shenglong and Gao, Zhangwei and Cui, Erfei and Tong, Wenwen and Hu, Kongzhi and Luo, Jiapeng and Ma, Zheng and others},
  journal={Science China Information Sciences},
  volume={67},
  number={12},
  pages={220101},
  year={2024},
  publisher={Springer}
}

@article{zhu2025internvl3,
  title={Internvl3: Exploring advanced training and test-time recipes for open-source multimodal models},
  author={Zhu, Jinguo and Wang, Weiyun and Chen, Zhe and Liu, Zhaoyang and Ye, Shenglong and Gu, Lixin and Tian, Hao and Duan, Yuchen and Su, Weijie and Shao, Jie and others},
  journal={arXiv preprint arXiv:2504.10479},
  year={2025}
}

@article{deng2025bagel,
  title   = {Emerging Properties in Unified Multimodal Pretraining},
  author  = {Deng, Chaorui and Zhu, Deyao and Li, Kunchang and Gou, Chenhui and Li, Feng and Wang, Zeyu and Zhong, Shu and Yu, Weihao and Nie, Xiaonan and Song, Ziang and Shi, Guang and Fan, Haoqi},
  journal = {arXiv preprint arXiv:2505.14683},
  year    = {2025}
}

@inproceedings{wang2025vggt,
  title={{VGGT}: Visual geometry grounded transformer},
  author={Wang, Jianyuan and Chen, Minghao and Karaev, Nikita and Vedaldi, Andrea and Rupprecht, Christian and Novotny, David},
  booktitle={Proceedings of the IEEE/CVF Conference on Computer Vision and Pattern Recognition,},
  pages={5294--5306},
  year={2025}
}

@article{wang2025pi3,
  title={$\pi^3$: Scalable Permutation-Equivariant Visual Geometry Learning},
  author={Wang, Yifan and Zhou, Jianjun and Zhu, Haoyi and Chang, Wenzheng and Zhou, Yang and Li, Zizun and Chen, Junyi and Pang, Jiangmiao and Shen, Chunhua and He, Tong},
  journal={arXiv preprint arXiv:2507.13347},
  year={2025}
}

@article{yang2025cambrian,
  title={Cambrian-s: Towards spatial supersensing in video},
  author={Yang, Shusheng and Yang, Jihan and Huang, Pinzhi and Brown, Ellis and Yang, Zihao and Yu, Yue and Tong, Shengbang and Zheng, Zihan and Xu, Yifan and Wang, Muhan and others},
  journal={arXiv preprint arXiv:2511.04670},
  year={2025}
}

@article{li2025spatialladder,
  title={Spatialladder: Progressive training for spatial reasoning in vision-language models},
  author={Li, Hongxing and Li, Dingming and Wang, Zixuan and Yan, Yuchen and Wu, Hang and Zhang, Wenqi and Shen, Yongliang and Lu, Weiming and Xiao, Jun and Zhuang, Yueting},
  journal={arXiv preprint arXiv:2510.08531},
  year={2025}
}

@article{vst2025,
  title={Visual Spatial Tuning},
  author={Yang, Rui and Zhu, Ziyu and Li, Yanwei and Huang, Jingjia and Yan, Shen and Zhou, Siyuan and Liu, Zhe and Li, Xiangtai and Li, Shuangye and Wang, Wenqian and Lin, Yi and Zhao, Hengshuang},
  journal={arXiv preprint arXiv:2511.05491},
  year={2025}
}

@article{wang2025ross3d,
  title={Ross3d: Reconstructive visual instruction tuning with 3d-awareness},
  author={Wang, Haochen and Zhao, Yucheng and Wang, Tiancai and Fan, Haoqiang and Zhang, Xiangyu and Zhang, Zhaoxiang},
  journal={arXiv preprint arXiv:2504.01901},
  year={2025}
}

@article{hu2025g2vlm,
  title={G$^2$ {VLM}: Geometry Grounded Vision Language Model with Unified 3D Reconstruction and Spatial Reasoning},
  author={Hu, Wenbo and Lin, Jingli and Long, Yilin and Ran, Yunlong and Jiang, Lihan and Wang, Yifan and Zhu, Chenming and Xu, Runsen and Wang, Tai and Pang, Jiangmiao},
  journal={arXiv preprint arXiv:2511.21688},
  year={2025}
}

@inproceedings{huang20253drs,
  title={3drs: Mllms need 3d-aware representation supervision for scene understanding},
  author={Huang, Xiaohu and Wu, Jingjing and Xie, Qunyi and Han, Kai},
  booktitle={The Thirty-ninth Annual Conference on Neural Information Processing Systems},
  year={2025}
}

@article{li2025spatialforcing,
  title={Spatial Forcing: Implicit Spatial Representation Alignment for Vision-language-action Model},
  author={Li, Fuhao and Song, Wenxuan and Zhao, Han and Wang, Jingbo and Ding, Pengxiang and Wang, Donglin and Zeng, Long and Li, Haoang},
  journal={arXiv preprint arXiv:2510.12276},
  year={2025}
}

@article{li2025recogdrive,
  title={Recogdrive: A reinforced cognitive framework for end-to-end autonomous driving},
  author={Li, Yongkang and Xiong, Kaixin and Guo, Xiangyu and Li, Fang and Yan, Sixu and Xu, Gangwei and Zhou, Lijun and Chen, Long and Sun, Haiyang and Wang, Bing and others},
  journal={arXiv preprint arXiv:2506.08052},
  year={2025}
}

@article{zhou2025roborefer,
  title={RoboRefer: Towards Spatial Referring with Reasoning in Vision-Language Models for Robotics},
  author={Zhou, Enshen and An, Jingkun and Chi, Cheng and Han, Yi and Rong, Shanyu and Zhang, Chi and Wang, Pengwei and Wang, Zhongyuan and Huang, Tiejun and Sheng, Lu and others},
  journal={arXiv preprint arXiv:2506.04308},
  year={2025}
}

@article{zhang2024video,
  title={Video instruction tuning with synthetic data},
  author={Zhang, Yuanhan and Wu, Jinming and Li, Wei and Li, Bo and Ma, Zejun and Liu, Ziwei and Li, Chunyuan},
  journal={arXiv preprint arXiv:2410.02713},
  year={2024}
}

@article{dauner2024navsim,
  title={Navsim: Data-driven non-reactive autonomous vehicle simulation and benchmarking},
  author={Dauner, Daniel and Hallgarten, Marcel and Li, Tianyu and Weng, Xinshuo and Huang, Zhiyu and Yang, Zetong and Li, Hongyang and Gilitschenski, Igor and Ivanovic, Boris and Pavone, Marco and others},
  journal={Advances in Neural Information Processing Systems},
  volume={37},
  pages={28706--28719},
  year={2024}
}

@article{cao2024physgame,
  title={Physgame: Uncovering physical commonsense violations in gameplay videos},
  author={Cao, Meng and Tang, Haoran and Zhao, Haoze and Guo, Hangyu and Liu, Jiaheng and Zhang, Ge and Liu, Ruyang and Sun, Qiang and Reid, Ian and Liang, Xiaodan},
  journal={arXiv preprint arXiv:2412.01800},
  year={2024}
}

@article{li2025does,
  title={Does Your 3D Encoder Really Work? When Pretrain-SFT from 2D VLMs Meets 3D VLMs},
  author={Li, Haoyuan and Zhou, Yanpeng and Gao, Yufei and Tang, Tao and Han, Jianhua and Yuan, Yujie and Chen, Dave Zhenyu and Bian, Jiawang and Xu, Hang and Liang, Xiaodan},
  journal={arXiv preprint arXiv:2506.05318},
  year={2025}
}

@inproceedings{fu2024blink,
  title={Blink: Multimodal large language models can see but not perceive},
  author={Fu, Xingyu and Hu, Yushi and Li, Bangzheng and Feng, Yu and Wang, Haoyu and Lin, Xudong and Roth, Dan and Smith, Noah A and Ma, Wei-Chiu and Krishna, Ranjay},
  booktitle={European Conference on Computer Vision},
  pages={148--166},
  year={2024},
  organization={Springer}
}

@inproceedings{du2024embspatial,
  title={Embspatial-bench: Benchmarking spatial understanding for embodied tasks with large vision-language models},
  author={Du, Mengfei and Wu, Binhao and Li, Zejun and Huang, Xuan-Jing and Wei, Zhongyu},
  booktitle={Proceedings of the 62nd Annual Meeting of the Association for Computational Linguistics (Volume 2: Short Papers)},
  pages={346--355},
  year={2024}
}

@article{liu2024ppllava,
  title={Ppllava: Varied video sequence understanding with prompt guidance},
  author={Liu, Ruyang and Tang, Haoran and Liu, Haibo and Ge, Yixiao and Shan, Ying and Li, Chen and Yang, Jiankun},
  journal={arXiv preprint arXiv:2411.02327},
  year={2024}
}

@inproceedings{jin2024chat,
  title={Chat-univi: Unified visual representation empowers large language models with image and video understanding},
  author={Jin, Peng and Takanobu, Ryuichi and Zhang, Wancai and Cao, Xiaochun and Yuan, Li},
  booktitle={Proceedings of the IEEE/CVF Conference on Computer Vision and Pattern Recognition},
  pages={13700--13710},
  year={2024}
}

@inproceedings{liu2024st,
  title={St-llm: Large language models are effective temporal learners},
  author={Liu, Ruyang and Li, Chen and Tang, Haoran and Ge, Yixiao and Shan, Ying and Li, Ge},
  booktitle={European Conference on Computer Vision},
  pages={1--18},
  year={2024},
  organization={Springer}
}

@inproceedings{li2024mvbench,
  title={Mvbench: A comprehensive multi-modal video understanding benchmark},
  author={Li, Kunchang and Wang, Yali and He, Yinan and Li, Yizhuo and Wang, Yi and Liu, Yi and Wang, Zun and Xu, Jilan and Chen, Guo and Luo, Ping and others},
  booktitle={Proceedings of the IEEE/CVF Conference on Computer Vision and Pattern Recognition},
  pages={22195--22206},
  year={2024}
}

@inproceedings{lin2024video,
  title={Video-llava: Learning united visual representation by alignment before projection},
  author={Lin, Bin and Ye, Yang and Zhu, Bin and Cui, Jiaxi and Ning, Munan and Jin, Peng and Yuan, Li},
  booktitle={Proceedings of the 2024 conference on empirical methods in natural language processing},
  pages={5971--5984},
  year={2024}
}

@article{bai2023qwen,
  title={Qwen-vl: A frontier large vision-language model with versatile abilities},
  author={Bai, Jinze and Bai, Shuai and Yang, Shusheng and Wang, Shijie and Tan, Sinan and Wang, Peng and Lin, Junyang and Zhou, Chang and Zhou, Jingren},
  journal={arXiv preprint arXiv:2308.12966},
  volume={1},
  number={2},
  pages={3},
  year={2023}
}

@article{achiam2023gpt,
  title={Gpt-4 technical report},
  author={Achiam, Josh and Adler, Steven and Agarwal, Sandhini and Ahmad, Lama and Akkaya, Ilge and Aleman, Florencia Leoni and Almeida, Diogo and Altenschmidt, Janko and Altman, Sam and Anadkat, Shyamal and others},
  journal={arXiv preprint arXiv:2303.08774},
  year={2023}
}

@misc{anthropic2024claude,
  author    = {Anthropic},
  title     = {Claude 3.5 sonnet},
  year      = {2024},
  url       = {https://www.anthropic.com/news/claude-3-5-sonnet}
}

@inproceedings{xu20253d,
  title={3d-more: Unified modal-contextual reasoning for embodied question answering},
  author={Xu, Rongtao and Gao, Han and Yu, Mingming and An, Dong and Chen, Shunpeng and Wang, Changwei and Guo, Li and Liang, Xiaodan and Xu, Shibiao},
  booktitle={2025 IEEE/RSJ International Conference on Intelligent Robots and Systems (IROS)},
  pages={5924--5929},
  year={2025},
  organization={IEEE}
}

@article{xu2025a0,
  title={A0: An affordance-aware hierarchical model for general robotic manipulation},
  author={Xu, Rongtao and Zhang, Jian and Guo, Minghao and Wen, Youpeng and Yang, Haoting and Lin, Min and Huang, Jianzheng and Li, Zhe and Zhang, Kaidong and Wang, Liqiong and others},
  journal={arXiv preprint arXiv:2504.12636},
  year={2025}
}

@article{zhang2024navid,
  title={NaVid: Video-based VLM Plans the Next Step for Vision-and-Language Navigation},
  author={Zhang, Jiazhao and Wang, Kunyu and Xu, Rongtao and Zhou, Gengze and Hong, Yicong and Fang, Xiaomeng and Wu, Qi and Zhang, Zhizheng and He, Wang},
  journal={arXiv preprint arXiv:2402.15852},
  year={2024}
}

@article{cai2025holistic,
  title={Holistic Evaluation of Multimodal LLMs on Spatial Intelligence},
  author={Cai, Zhongang and Wang, Yubo and Sun, Qingping and Wang, Ruisi and Gu, Chenyang and Yin, Wanqi and Lin, Zhiqian and Yang, Zhitao and Wei, Chen and Qian, Oscar and others},
  journal={arXiv preprint arXiv:2508.13142},
  year={2025}
}
